\documentclass[letterpaper,twocolumn,10pt]{article}
\usepackage{usenix-2020-09}
\usepackage{graphicx}
\usepackage{tcolorbox}
\usepackage{booktabs}      
\usepackage{multirow}      
\usepackage{xcolor}        
\usepackage{pifont}        
\usepackage{xspace}        
\usepackage{marvosym}

\definecolor{ForestGreen}{RGB}{34,139,34}
\definecolor{orange}{RGB}{255,140,0}
\definecolor{darkgreen}{RGB}{0,130,0}        
\definecolor{darkred}{RGB}{220,20,20}        
\newcommand{\systemname}{\textsc{MobiMem}\xspace}
\newcommand{\FEH}[1]{\textcolor{black}{#1}}

\newenvironment{myitemize}%
  {\begin{list}{\labelitemi}{\itemsep1pt \topsep2pt \parsep0.00in
  \partopsep=0pt \leftmargin1em}}%
  {\end{list}}

\begin{document}
\begin{sloppypar}
	\title{Beyond Training: Enabling Self-Evolution of Agents with \systemname}

	\author{Zibin Liu$^{1\dagger}$, Cheng Zhang$^{1\dagger}$, Xi Zhao$^{1\dagger}$, Yunfei Feng$^{\dagger}$, Bingyu Bai$^{\dagger}$,
	\\ Dahu Feng$^{\dagger}$, Erhu Feng\textsuperscript{\ding{41}}$^{\dagger}$, Yubin Xia$^{\dagger}$, Haibo Chen $^{\dagger}$
	\\
	{\normalsize \it
	{$^\dagger$Institute of Parallel and Distributed Systems, Shanghai Jiao Tong University}} \\
	}

	\maketitle

	\footnotetext[1]{The three authors contributed equally to this work and should be considered co-first authors.}
	{\renewcommand{\thefootnote}{\ding{41}}
	\footnotetext[2]{Erhu Feng is the corresponding author: fengerhu1@sjtu.edu.cn}}
	
	\begin{abstract}
    Large Language Model (LLM) agents are increasingly deployed to automate complex workflows in mobile and desktop environments.
    However, current \emph{model-centric} agent architectures struggle to self-evolve post-deployment:
    improving personalization, capability, and efficiency typically requires continuous model retraining/fine-tuning,
    which incurs prohibitive computational overheads and suffers from an inherent trade-off between model accuracy and inference efficiency.

    To enable iterative self-evolution without model retraining, we propose \systemname,
    a \emph{memory-centric} agent system.
    \systemname first introduces three specialized memory primitives to decouple agent evolution from model weights:
    (1) \emph{Profile Memory} uses a lightweight distance-graph (DisGraph) structure to align with user preferences, resolving the accuracy-latency trade-off in user profile retrieval; 
    (2) \emph{Experience Memory} employs multi-level templates to instantiate execution logic for new tasks, ensuring capability generalization; and 
    (3) \emph{Action Memory} records fine-grained interaction sequences, reducing the reliance on expensive model inference.
    Building upon this memory architecture, \systemname further integrates a suite of OS-inspired services to orchestrate execution: 
    a scheduler that coordinates parallel sub-task execution and memory operations;
    an agent record-and-replay (AgentRR) mechanism that enables safe and efficient action reuse; 
    and a context-aware exception handling that ensures graceful recovery from user interruptions and runtime errors.

    Evaluation on AndroidWorld and top-50 apps shows that \systemname achieves 83.1\% profile alignment
    with 23.83 ms retrieval time (280$\times$ faster than GraphRAG baselines), 
    improves task success rates by up to 50.3\%,
    and reduces end-to-end latency by up to 9$\times$ on mobile devices,
    demonstrating the efficiency and practicality of memory-centric evolution in real-world deployments.

\end{abstract}
	\section{Introduction}

The rapid advancement of large language models (LLMs) has catalyzed the emergence and proliferation of AI agents, 
which are capable of autonomously executing complex tasks through natural language understanding and tool manipulation.
As AI agents~\cite{ye2025mobileagent, wang2025uitars2, pumariega2025agents3} evolve, they are increasingly deployed in mobile and desktop environments to automate user workflows~\cite{zapier_workflow, microsoft_copilot_agents, ibm_agentic_workflows, power_automate}, spanning from simple information retrieval to multi-step, cross-application tasks that require coordinating operations across diverse apps and services.
For example, agents may help users book hotels by comparing prices across multiple travel apps, 
schedule meetings by coordinating calendar and communication apps, etc.

\begin{figure}[t]
    \centering
    \setlength{\abovecaptionskip}{-1pt}
    \setlength{\belowcaptionskip}{-15pt}
    \includegraphics[width=\columnwidth]{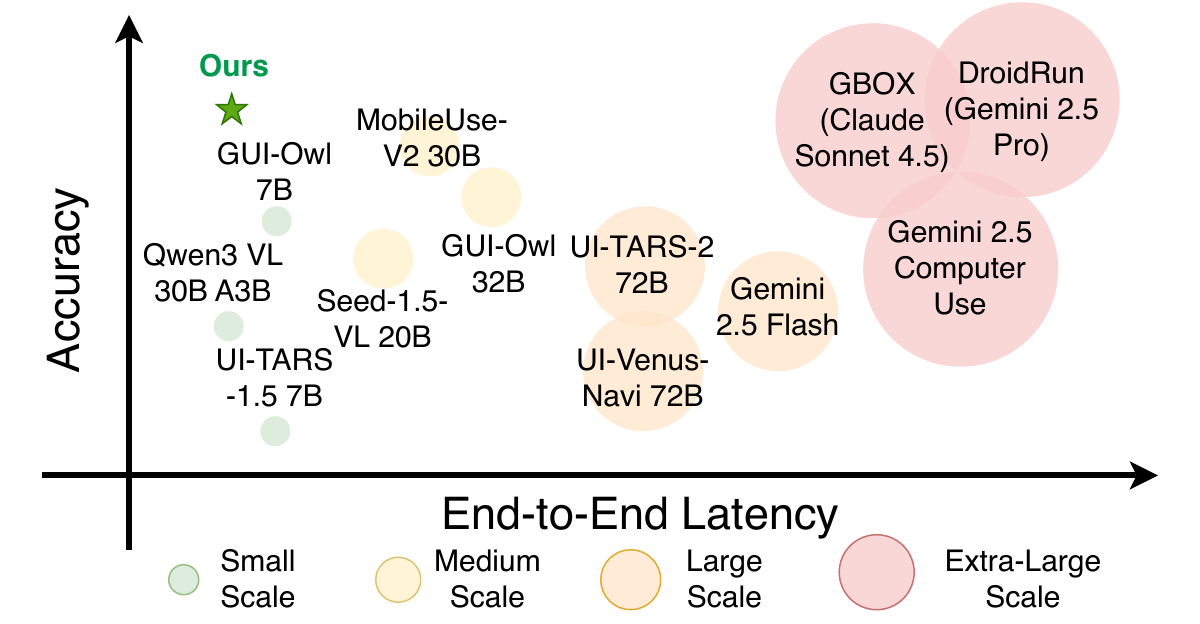}
    \caption{\systemname tames the trade-off between AI agents' latency and accuracy by a memory-centric design.}
    \label{fig:model-compare}
  \end{figure}

\FEH{Although contemporary AI agents demonstrate autonomous planning, decision-making, and execution capabilities, 
they often lack the capacity for continual evolution during deployment.
A self-evolving agent should aim to achieve three objectives: (1) continually learn user preferences, 
(2) continually expand its capabilities, and (3) continually improve its execution efficiency.
Given that current agents are predominantly built with \textbf{model-centric} architectures, 
such evolution is achieved through per-user model fine-tuning~\cite{hong2024cogagent,wu2024osatlas,lin2024showui,zhang2025agentcpm,Cheng2024SeeClick,hu2021lora} 
or reinforcement learning~\cite{verl,gu2025uivenus,xu2025mobilerl,dong2025agenticreinforcedpolicyoptimization,lu2025arpoendtoendpolicyoptimizationgui,feng2025groupingroup}, 
which incurs prohibitive computational costs for continual training both on-device and in the cloud.
Moreover, relying solely on model-based approaches makes it difficult to balance the agent's performance and accuracy, 
as shown in Figure~\ref{fig:model-compare}. 
For example, a powerful agent system commonly requires multimodal models with tens to hundreds of billions of parameters. 
Such models are not only difficult to deploy on edge devices but also introduce substantial latency in cloud deployment, 
thereby limiting their practical utility in real-world scenarios.}

\FEH{
To avoid these post-deployment training overheads, 
we propose a \textbf{memory-centric} agent system: \systemname, 
which supports agent self-evolution without the requirement for additional model training.
\systemname draws inspiration from classical OS mechanisms, 
including memory management, scheduling, record-and-replay, etc., and adapts them to agentic scenarios.
Through fine-grained memory updating, caching, and orchestration, 
\systemname supports continuous improvement in personalization, capability, and efficiency. }

\FEH{\systemname organizes agent memory into three types: Profile Memory, Experience Memory, and Action Memory.}

\begin{myitemize}
    \item \textbf{\emph{Profile Memory:}} \FEH{used to learn user preferences, 
    enabling the agent to continually personalize its behavior. 
    Although existing studies~\cite{packer2023memgpt,mem02025,wang2025mirix} have attempted to capture user preferences 
    through techniques such as RAG~\cite{rag1}, GraphRAG~\cite{graphrag}, and hierarchical memory~\cite{packer2023memgpt}, 
    they still face a trade-off between retrieval accuracy and efficiency. 
    RAG enables efficient information extraction through vector search but lacks explicit relationships between entities. 
    GraphRAG provides richer relational information, but retrieval and update require model inference, 
    introducing non-negligible latency.
    To address this, we design the DisGraph, which shifts semantic information from edges to nodes, 
    while edges solely encode relational distances between nodes. 
    It preserves profile accuracy while reducing the latency of retrieval and update operations.
    }
    \item \textbf{\emph{Experience Memory:}} \FEH{storing experience templates, 
    enabling the agent to continuously improve its capability. 
    Prior work~\cite{autodroid,AWM} only records the agent's execution trajectories as experience, 
    but this approach compromises the agent's generalization. 
    To address this limitation, \systemname introduces a multi-level, multi-step experience template. 
    When a new task arrives, \systemname inherits relevant templates and instantiates the execution steps required for the current task, 
    ensuring that the experience remains applicable across diverse scenarios.
    }
    \item \textbf{\emph{Action Memory:}} \FEH{recording the interactions between the agent and the system, 
    allowing the agent to continually improve its execution efficiency. 
    Similar to the procedural or ``muscle'' memory that humans develop through repeated practice, 
    the agent can directly replay the action sequences stored in action memory for common tasks, 
    without relying on model-based reasoning.
    To address issues analogous to cache staleness, 
    \systemname introduces a lightweight mechanism to determine whether an action memory entry is a valid hit (reusable) 
    or to handle cases where the action memory has become stale.}
\end{myitemize}

\FEH{With the support of agent memory, 
\systemname further provides system-level agent services, 
including the fine-grained task scheduler, agent-level record-and-replay mechanism, 
agent interrupt and exception handler, etc. 
Leveraging the experience memory, 
\systemname establishes fine-grained dependencies at the subtask level, 
enabling independent subtasks to execute in parallel. 
Through the action memory, \systemname offers an agent-level record-and-replay capability (AgentRR). 
Unlike system-level record-and-replay services~\cite{revirt, r2, odr, laadan2010transparent}, 
AgentRR does not rigidly reproduce previous actions; 
instead, it preserves the agent's generalization ability during replay. 
For interruption and exception handling, 
\systemname maintains the context of the agent's execution process 
and integrates the corresponding exception handlers into its experience memory, 
ensuring that subsequent executions are not disrupted by the same exception.}

We evaluate our system through extensive experiments on AndroidWorld~\cite{rawles2024androidworld} benchmark 
and real-world workloads (including top 50 mobile Apps) across diverse hardware platforms.
Profile Memory achieves 83.1\% profile alignment with 23.83 ms retrieval latency, 
outperforming baseline approaches by up to 25\% and achieving over 280$\times$ speedup compared to GraphRAG methods.
Experience Memory improves task success rates by up to 50.3\% across four agent models, with near-zero human effort through automated abstraction.
Action Memory achieves 77.3\% average action reuse rate with human-crafted templates, 
reducing end-to-end latency by up to 9$\times$ on resource-constrained mobile devices.
In multi-task scenarios, our fine-grained task scheduling achieves up to 1.98$\times$ speedup 
by exploiting parallelism across independent sub-tasks.
Moreover, Experience Memory and AgentRR technologies have already been deployed in a flagship smartphone.

	\nopagebreak
	\section{Background}

\newcommand{\xmark}{\ding{55}}

\begin{table*}[t]
    \centering
    \small
    \setlength{\abovecaptionskip}{0pt}
    \setlength{\belowcaptionskip}{2pt} 
    \caption{Comparison of GUI agent frameworks and memory systems across memory architecture and system capabilities.}
    \label{tab:related_work}
    \begin{tabular}{l@{\hspace{12pt}}c@{\hspace{12pt}}c@{\hspace{12pt}}c@{\hspace{16pt}}c@{\hspace{12pt}}c@{\hspace{12pt}}c}
    \hline\hline
    & \multicolumn{3}{c}{\textbf{Memory Architecture}} & \multicolumn{3}{c}{\textbf{System Capabilities}} \\
    \cmidrule(lr){2-4} \cmidrule(lr){5-7}
    \textbf{System} & \textbf{User Profile} & \textbf{Execution} & \textbf{Action} & \textbf{Task} & \textbf{Error} & \textbf{User} \\
     & \textbf{Memory} & \textbf{Memory} & \textbf{Memory} & \textbf{Scheduling} & \textbf{Recovery} & \textbf{Interrupt} \\
    \midrule
    UI-TARS~\cite{ui-tars} & \textcolor{darkred}{\xmark} & \textcolor{orange}{Training-time data} & \textcolor{darkred}{\xmark} & \textcolor{orange}{Sequential} & \textcolor{darkred}{\xmark} & \textcolor{darkred}{\xmark} \\
    MA3~\cite{ye2025mobileagent} & \textcolor{orange}{RAG module} & \textcolor{orange}{Compressed histories} & \textcolor{darkred}{\xmark} & \textcolor{orange}{Sequential} & \textcolor{orange}{LLM Reflector} & \textcolor{darkred}{\xmark} \\
    AutoDroid~\cite{autodroid} & \textcolor{darkred}{\xmark} & \textcolor{orange}{Simulated traces} & \textcolor{orange}{Task-level} & \textcolor{orange}{Sequential} & \textcolor{darkred}{\xmark} & \textcolor{darkred}{\xmark} \\
    \midrule
    MemGPT~\cite{packer2023memgpt} & \textcolor{orange}{Profile blocks (Key-value)} & \textcolor{darkred}{\xmark} & \textcolor{darkred}{\xmark} & - & - & - \\
    Mem0~\cite{mem02025} & \textcolor{orange}{User facts (Graph)} & \textcolor{darkred}{\xmark} & \textcolor{darkred}{\xmark} & - & - & - \\
    A-MEM~\cite{xu2025amem} & \textcolor{orange}{General notes (Graph)} & \textcolor{darkred}{\xmark} & \textcolor{darkred}{\xmark} & - & - & - \\
    \midrule
    \textbf{Our Work} & \textbf{\textcolor{darkgreen}{Concept-Entity}} & \textbf{\textcolor{darkgreen}{Experience}} & \textbf{\textcolor{darkgreen}{Tree/Chain}} & \textbf{\textcolor{darkgreen}{Fine-grained}} & \textbf{\textcolor{darkgreen}{Iterative}} & \textbf{\textcolor{darkgreen}{Exception}} \\
     & \textbf{\textcolor{darkgreen}{(DisGraph)}} & \textbf{\textcolor{darkgreen}{Templates}} & \textbf{\textcolor{darkgreen}{Memory}} & \textbf{\textcolor{darkgreen}{Parallelism}} & \textbf{\textcolor{darkgreen}{Refinement}} & \textbf{\textcolor{darkgreen}{Handling}} \\
    \hline\hline
    \end{tabular}\\[-5pt]
    \vspace{-0.2cm}
\end{table*}

\subsection{Mobile Agent Frameworks}
Recent advances in reasoning paradigms~\cite{yao2023react,zhang2024coat} and GUI parsing tools~\cite{yu2025omniparser} enable GUI agent frameworks to adopt iterative execution paradigms where agents perceive screenshots, generate reasoning and actions autonomously.
UI-TARS~\cite{ui-tars} implements an online trace bootstrapping framework where the model dynamically learns from past task executions through continual training.
Mobile-Agent-v3(MA3)~\cite{ye2025mobileagent} introduces GUI-Owl as a foundational model.
The framework maintains a compressed history storing interaction traces as the execution memory of the ongoing task.
It also leverages an RAG module storing external or user-specific knowledge and a Reflector Agent for error diagnosis and recovery.
AutoDroid~\cite{autodroid} employs an App Memory module built in an offline stage to store simulated task traces which will be used to synthesize execution guidelines, or be fully reused for similar tasks during online deployment.
Related frameworks explore variants with hierarchical planning~\cite{mobile-agent-e}, shortcut learning~\cite{appagentx}, lifelong learning~\cite{voyager}, or broader OS-level scope~\cite{ufo, os-copilot}.

While these frameworks demonstrate advances in execution memory and error handling (Table~\ref{tab:related_work}), they share fundamental limitations in memory architecture and system capabilities.
On the memory side, they lack structured user profile memory for personalization and fine-grained action memory to fully exploit shared actions, while their execution memories store raw traces rather than distilled templates.
On the system side, they employ sequential task scheduling without exploiting parallelism opportunities, and they do not support user interventions to correct agent errors during execution.
Our work addresses these gaps through a comprehensive memory architecture that captures user profiles, execution experiences, and reusable action patterns, coupled with system capabilities including fine-grained parallelism, iterative refinement for error recovery, and exception handling mechanisms that enable user intervention and correction.

\subsection{Memory Systems in AI Agents}
Memory in agent architectures can be categorized into short-term memory, which maintains immediate context within the model's input window, and long-term memory, which uses external storage to persist information across sessions~\cite{wu2025human,shan2025cognitive}.
Recent research focuses on enhancing long-term memory capabilities through mechanisms including indexing and retrieval~\cite{hipporag,locomo}, dynamic updating and forgetting~\cite{bae2022keep,memorybank}, and memory-enhanced generation~\cite{memorag}.
Building on these foundations, several memory systems enhance agent capabilities beyond single-task execution.
MemGPT~\cite{packer2023memgpt} introduces a hierarchical memory management system inspired by OS virtual memory, organizing information into main context, external context, and archival storage with self-editing capabilities through key-value structures.
Mem0~\cite{mem02025} focuses on automatic fact extraction and deduplication, merging conversational statements into a consolidated graph structure to maintain user-specific facts across interactions.
A-MEM~\cite{xu2025amem} emphasizes agentic self-organization of knowledge, where agents autonomously generate comprehensive memory notes with rich metadata and dynamically establish inter-memory links to form an evolving knowledge network.
Other systems explore diverse architectures including human-like memory~\cite{myagent}, memo-based mechanisms~\cite{memochat}, self-controlled frameworks~\cite{wang2024scm}, unified memory architectures~\cite{li2025memos}, and hybrid multimodal memory~\cite{li2024optimus}.

However, as shown in Table~\ref{tab:related_work}, these memory systems are designed for general conversational contexts rather than GUI automation domains.
They excel at storing declarative knowledge (user facts, general notes) but lack the specialized memory structures needed for GUI agents.
Specifically, they do not maintain execution memory to capture procedural knowledge from past interactions or provide action memory to cache reusable interaction patterns and templates.
Our work optimizes memory architecture for GUI agents through Concept-Entity graphs that enable efficient multi-dimensional user profile retrieval, experience templates that distill execution traces, and Action Memory structures that cache reusable action sequences.

	\nopagebreak
	\section{System Overview}
\label{sec:overview}

\begin{figure*}[t]
  \centering
  \setlength{\abovecaptionskip}{-1pt}
  \setlength{\belowcaptionskip}{-10pt}
  \includegraphics[width=\textwidth]{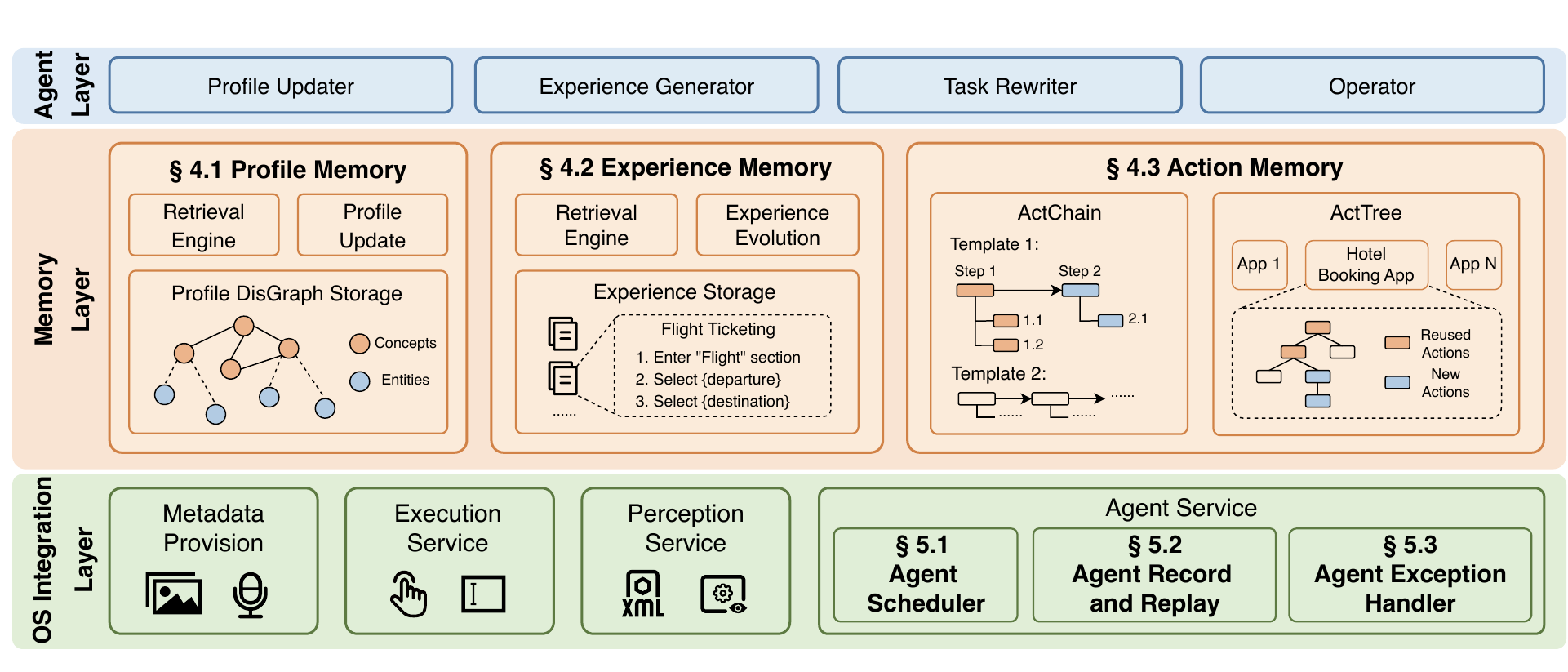}
  \caption{Three-layer architecture of \systemname: specialized multi-agent layer (top), 
  agent-tailored memory layer (middle), 
  and OS integration layer (bottom).}
  \label{fig:architecture}
\end{figure*}

\FEH{Existing AI-agent systems still face challenges in continually evolving during deployment, 
such as achieving agent personalization and improving agent capability and efficiency without continual model training.
To address these issues, we design \systemname (Figure~\ref{fig:architecture}) with three hierarchical layers: 
a multi-agent layer, an agent memory layer, and an OS integration layer that provides agent services and coordinates system components.
}

\textbf{Multi-agent Layer.}
Our system collaborates with four specialized agents that work together to execute user tasks.
The \textit{Profile Updater} processes OS metadata to extract and update user profiles into Profile Memory.
The \textit{Experience Generator} distills reusable task templates from execution histories and stores them in Experience Memory.
The \textit{Task Rewriter} matches incoming user requests to existing templates and fills template parameters for execution.
The \textit{Operator} generates actions when action reuse is not applicable or when encountering unrecorded situations.
These agents leverage our memory system to access persistent context and reusable actions, reducing redundant reasoning.

\textbf{Memory Layer.}
Our memory system offers three core modules (\S\ref{subsec:user-profile}-\ref{subsec:action-cache}) to provide personalization and improve task success rates and execution efficiency.
\FEH{\textit{Profile Memory Module} (\S\ref{subsec:user-profile}) addresses the lack of personalization by organizing user preferences, facts, 
and behavior patterns in a DisGraph structure, 
which requires only one LLM call for updates and zero-LLM retrievals through embedding-based search and graph traversal.}
\textit{Experience Memory Module} (\S\ref{subsec:experience-template}) distills shared execution patterns by decomposing task execution 
into invariant control logic and variable parameters and storing them as experience templates that the Task Rewriter Agent can utilize.
\FEH{\textit{Action Memory Module} (\S\ref{subsec:action-cache}) reduces LLM inference overhead by caching historical actions in two structures: 
ActTree for prefix reuse and ActChain for prefix-suffix reuse. 
\systemname also designs a fallback mechanism for Action Memory to handle cache-miss scenarios.}

\textbf{OS Integration Layer.}
\FEH{Our system is built on an OS with first-class agent support, providing three categories of agent service:
\textit{Agent Scheduler} (\S\ref{subsec:scheduler}) that orchestrates parallel sub-tasks according to the fine-grained experience memory, 
\textit{AgentRR} (\S\ref{subsec:agentrr}), a record-and-replay mechanism that captures execution traces for cache population and enables safe action reuse, 
and an \textit{Agent Exception Handler} (\S\ref{subsec:exception-handling}) that enables graceful recovery from user interruptions, 
and records each exception handler in the experience memory.
In addition to these agent services, \systemname also provides agents with system-level perception and execution runtime services, including:
\textit{metadata provision} (e.g., screenshots, voice recordings) enables agents to analyze and learn user habits;
\textit{perception service} (e.g., UI hierarchy, element visibility and semantics) helps agents comprehend system and application state;
\textit{execution service} provides programmatic OS-level APIs for agents to perform UI interactions and event monitoring.}
	\nopagebreak
	\section{Agent Memory}
\label{sec:design}

\FEH{This section presents the detailed designs of three types of agent memory: Profile Memory, Experience Memory, and Action Memory. 
We first describe how Profile Memory organizes a user's personalized information, including facts, preferences, and behavior patterns (\S\ref{subsec:user-profile}). 
Next, we introduce how Experience Memory extracts and applies generalizable task-execution templates at multiple levels (\S\ref{subsec:experience-template}). 
Finally, we explain how Action Memory accelerates task execution through action reuse (\S\ref{subsec:action-cache}).
}

\subsection{Profile Memory}
\label{subsec:user-profile}

\textbf{Challenges.}
\FEH{As AI agents interact with users over time, they gradually accumulate rich personalized information, 
including factual details, preferences, and habitual behaviors. 
To deliver truly personalized services, an agent must continuously interpret and retain the user's evolving preferences. 
Current agent systems attempt to support long-term memory through RAG-based techniques, 
but they still face inherent trade-offs between performance and accuracy.}
Simple RAG approaches offer fast retrieval but suffer from low accuracy, 
as embedding-based search often returns semantically similar but irrelevant information.
Graph-based RAG systems improve accuracy but incur prohibitive latency, 
requiring expensive LLM calls for both graph updates and query traversal.

\begin{figure}[t]
  \centering
  \setlength{\abovecaptionskip}{-1pt}
  \setlength{\belowcaptionskip}{-10pt}
  \includegraphics[width=\columnwidth]{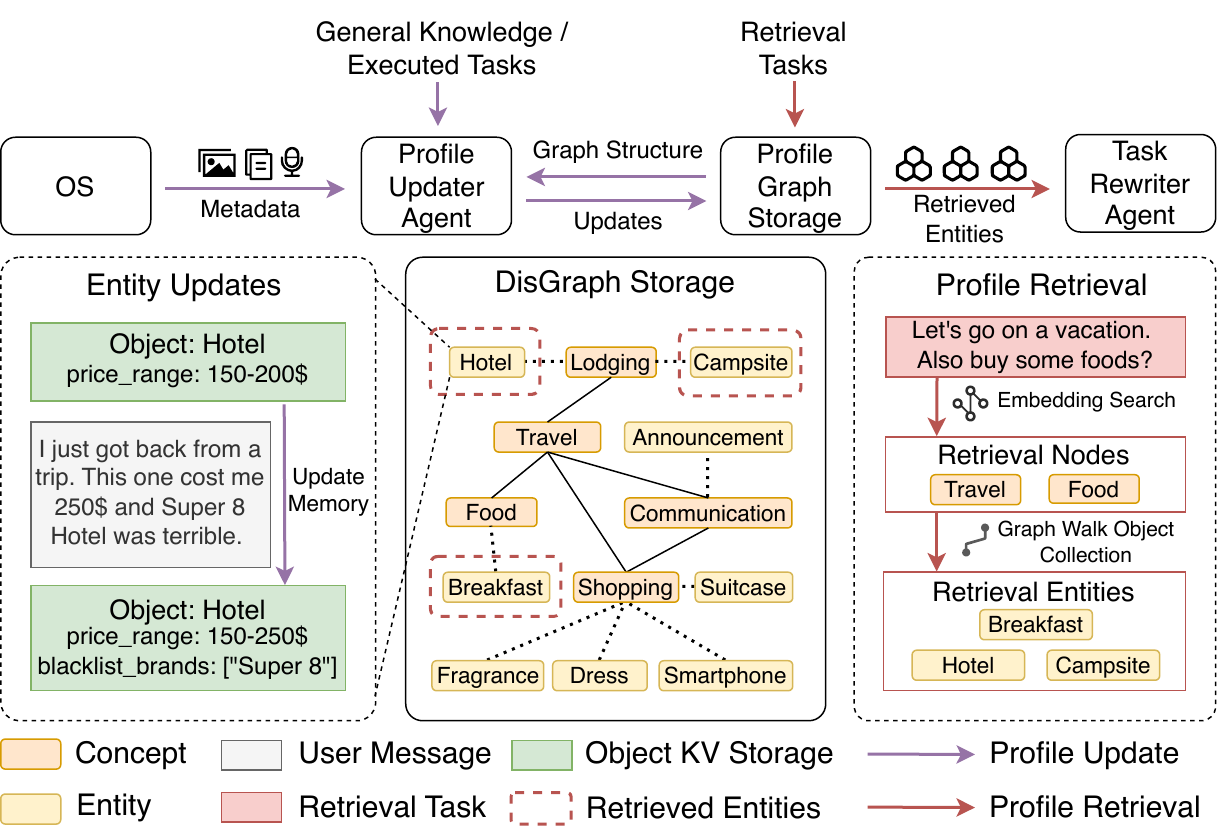}
  \caption{Architecture of the Profile Memory module, showing profile updating, storage, and retrieval workflows.}
  \label{fig:user-profile}
\end{figure}

\textbf{Key Insight.} 
Our analysis reveals that this trade-off stems from how structural relationships are represented.
Simple RAG's flat structure lacks relational context, causing embedding-based retrieval to fetch semantically similar but contextually inappropriate information.
GraphRAG improves accuracy by encoding rich relational semantics in edge weights, 
but this design necessitates expensive LLM calls to generate and traverse these semantic edges.
\FEH{Our key insight is to use a \textit{DisGraph} architecture that shifts semantic information from edges to nodes.}
The graph comprises two types of nodes: abstract concepts and concrete entities, 
where entities connect only to concepts and concepts interconnect with each other.
All edges in the graph are semantic-free, only indicating membership or relevance.
\FEH{Under this design, the relationship between any two nodes is correlated with the distance between them, 
with shorter paths generally indicating higher relevance.}
During retrieval, we only leverage embedding similarity to identify the most relevant starting nodes, 
then perform breadth-first search to expand to nodes within short path distances, gathering related profile information.
Therefore, the entire retrieval process does not require any LLM involvement.

\textbf{Our Design.}
As shown in Figure~\ref{fig:user-profile}, the Profile Memory module operates through two workflows: 
an update workflow where Profile Updater Agent processes OS metadata and task traces to extract the profile information into the DisGraph, 
and a retrieval workflow where incoming tasks trigger embedding-based search and BFS traversal to gather relevant context.
The storage organizes profile knowledge in the DisGraph architecture, with entities storing structured key-value pairs capturing user information.
\FEH{Operations on DisGraph mainly include three primitives: updating, retrieval, and splitting.}

\textit{Update logic.}
When new observations arrive, the system first performs embedding-based retrieval to identify relevant nodes in the DisGraph.
It then sends these retrieved nodes along with their outgoing edges and the new observations to the Profile Updater in a single LLM call.
The agent outputs structured modifications (updates to existing entities or insertions of new entities), applied atomically to evolve the profile.
\FEH{Figure \ref{fig:user-profile} illustrates how DisGraph updates the attributes of entities after obtaining a user's hotel reviews.}

\textit{Retrieval logic.}
When a new task arrives, the system uses an embedding model to score the task description against all entities and concepts in the graph, 
selecting the top-k most relevant nodes as starting points.
The system then performs breadth-first search from these starting points to expand the context.
\FEH{To balance coverage and relevance, the system partitions the profile context window into $k$ equal shares, 
and uses round-robin scheduling to fill each share with content discovered from each starting point.
Figure \ref{fig:user-profile} illustrates an example in which the task ``Let's go on a vacation this weekend. Also buy some foods?'' 
triggers an embedding search that identifies Travel and Food as the starting nodes and subsequently collects content related to these entities.}

\textit{Dynamic splitting.}
\FEH{As the user profile expands, the number of entity nodes linked to a popular concept node increases, 
which in turn reduces retrieval precision. 
The BFS expansion retrieves an excessive number of entities under the same concept, which dilutes the effective context window.}
To address this, when a concept's entity count exceeds a threshold, the system invokes the Profile Updater to analyze entity attributes and create specialized subconcepts 
(e.g., splitting Travel into Business Travel and Leisure Travel), then redistributes entities to their most relevant subconcepts.

\subsection{Experience Memory}
\label{subsec:experience-template}

\begin{figure}[t]
  \centering
  \setlength{\abovecaptionskip}{-1pt}
  \setlength{\belowcaptionskip}{-10pt}
  \includegraphics[width=\columnwidth]{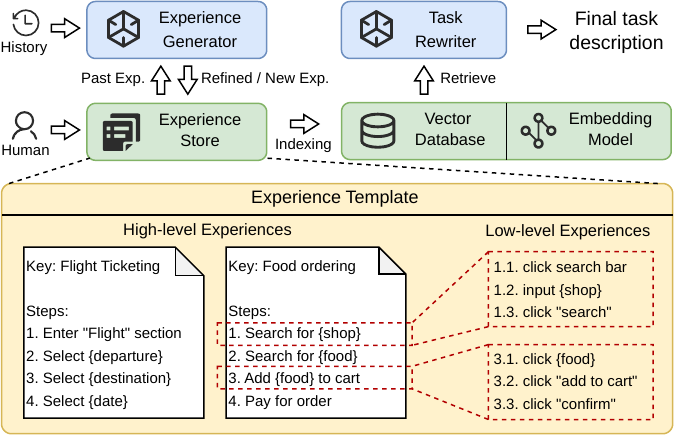}
  \caption{Experience template example showing template structure and experience generation/retrieval workflows.}
  \label{fig:template-example}
\end{figure}

\textbf{Challenges.}
\FEH{Current approaches enhance agent capabilities primarily through training on domain-specific data. 
However, such training-based approaches face inherent scalability challenges in real-world deployment.
They require large volumes of high-quality execution traces, which are costly to collect, 
leaving most long-tail tasks insufficiently covered. 
In addition, model training demands substantial computational resources. 
On-device training on endpoints such as smartphones remains impractical, 
while performing cloud-based training for each user's long-tail tasks also incurs significant cost.
}

\textbf{Key Insight.}
\FEH{Rather than training the model to understand complex pages, learn page-transition relationships, and perform task planning, 
it is more effective to directly provide the model with correct execution experiences, 
thereby reducing its reasoning burden. 
However, due to the complexity of task descriptions and environments, 
providing experiences for every task-environment combination is also not scalable.
To address this, we propose the \emph{multi-level experience} and \emph{experience template}. 
For experience templates, we observe that similar tasks share invariant control flow but have variant data flow.
This enables us to abstract execution patterns into templates with parameter slots.
For any concrete task, the agent inherits the corresponding template and instantiates parameter slots 
with task-specific values to produce an executable experience.
Moreover, multi-level experience strikes a balance between generality and specificity. 
High-level experience generalizes to a wider range of scenarios but demands greater capability from the model, 
whereas low-level experience presents the opposite trade-off.}

\textbf{Our Design.}
Based on this insight, we design an Experience Memory module that enables agents to accumulate, abstract, and reapply execution knowledge.
It consists of two core components: (1) an \textit{Experience Store} that maintains multi-level experience templates, 
and (2) a \textit{Vector Database} that enables fast semantic retrieval.
\FEH{As shown in Figure~\ref{fig:template-example}, higher-level experiences describe task-level control flow, 
while lower-level experiences provide concrete execution steps for precise navigation and interaction (e.g., click, swipe, etc.).
More specifically, the Experience Memory module provides two core mechanisms:}

\emph{Experience generation and storage.}
Experience templates are automatically synthesized when encountering new task classes, 
or manually authored by developers to bootstrap common tasks.
For automatic synthesis, when a task cannot find a matching template during retrieval, 
the Experience Generator synthesizes a new template by referencing similar past experiences.
For manual authoring, developers can directly author parameterized workflows.
All templates are keyed by their core descriptions, indexed using an embedding model and stored in the vector database for fast retrieval.

\emph{Template retrieval and parameter filling.}
When a new task arrives, the system queries the vector database to retrieve the best-matching template based on the semantic similarity 
between the task description and existing template keys.
The retrieved template is passed to the Task Rewriter to fill the parameter slots using information extracted from the current task.
\FEH{If no suitable experience template is found, the agent autonomously decides how to execute the task. 
Once the task is completed successfully, the Experience Generator will attempt to create a new experience template for future use.}

\textit{Cross-app task support.}
In real-world scenarios, a single task may span multiple applications (e.g., compare prices across shopping apps).
The Experience Memory module handles such cases via a DAG-based orchestration: 
experiences are modeled as a DAG of subtasks executed in topological order.
\FEH{Each subtask can declare parameter slots whose concrete values are determined by the outputs of its preceding subtasks.
More details will be introduced in \S\ref{subsec:scheduler}}.

\subsection{Action Memory}
\label{subsec:action-cache}

\begin{figure}[t]
  \centering
  \setlength{\abovecaptionskip}{-1pt}
  \setlength{\belowcaptionskip}{-15pt}
  \includegraphics[width=\columnwidth]{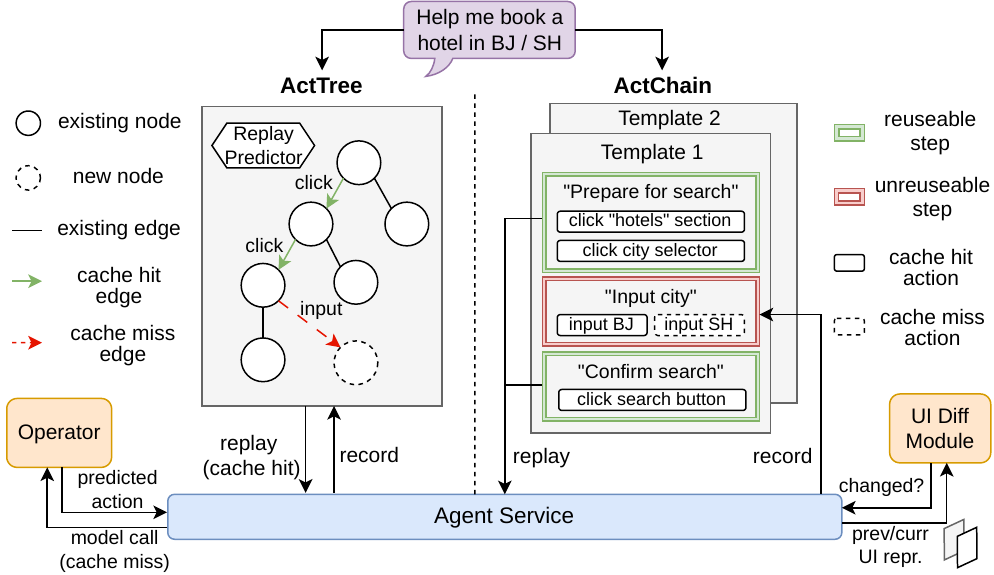}
  \caption{\FEH{Two structures of Action Memory: ActTree (left) for prefix reuse and ActChain (right) for prefix-suffix reuse.}}
  \label{fig:action-cache}
\end{figure}

\textbf{Challenges.}
Pure LLM-based agents lack the inherent ability to self-improve execution efficiency after deployment.
Even when executing previously completed tasks, 
these agents repeat the same reasoning process due to their stateless nature, 
resulting in consistently high latency, particularly for long-horizon tasks.
Existing systems such as AutoDroid have attempted to address this through task-level action caching, 
which reuses entire execution traces for similar tasks.
However, this approach fails to generalize across tasks with partially shared sub-procedures, 
and lacks mechanisms to detect environment changes, 
causing low cache hit rates and incorrect trace reuse in non-deterministic environments.

\textbf{Key Insight.}
\FEH{Our key insight is that, when humans perform similar or repetitive tasks, 
they often operate subconsciously and do not engage in deliberate, complex reasoning.
In agentic scenarios, we also observe that tasks within the same application typically share common prefixes; 
consequently, these actions can be directly reused without invoking the model. 
By organizing execution traces into a tree structure, we can identify shared prefixes at runtime using a lightweight embedding model, 
analogous to human procedural memory.}
For tasks bound to the same experience, we can achieve even more aggressive optimization.
Drawing from the analysis in Section~\ref{subsec:experience-template}, we can decompose a task into a sequence of invariant and variable steps.
Invariant steps remain unchanged across executions and can be directly reused, 
while variable steps are dependent on specific parameters and can only be reused 
if their parameters exactly match those of previously executed steps.
This decomposition allows us to cache and reuse the stable parts of an agent workflow without model reasoning.

\textbf{Our Design.}
Based on the above idea, the Action Memory module organizes interactions between the agent and system in two structures to support fine-grained action-level reuse: 
\FEH{\textit{ActTree} for prefix reuse and \textit{ActChain} for prefix-suffix reuse, as shown in Figure~\ref{fig:action-cache}.}

\textit{Prefix reuse.}
When the incoming task is not bound to any experience template, the Action Memory module works in the prefix reuse mode.
\FEH{This assumes that tasks within the same app can typically reuse prefix actions and share prefix pages.
However, once an action branch occurs, the consistency of the subsequent environment can no longer be guaranteed.}
For each application, it maintains an app-level memory with an ActTree structure where nodes represent UI states 
and edges represent state transitions with corresponding actions, 
aggregating all action sequences from previously completed tasks.
\FEH{We design a lightweight embedding model as a Replay Predictor to determine 
whether the current task can reuse actions completed by the historical task at the $n$-th layer of ActTree. 
The Action Memory module will adaptively select the reuse threshold to enhance reuse accuracy.}

\textit{Prefix-suffix reuse.}
\FEH{When an incoming task is associated with an experience template, 
the Action Memory module switches to a prefix-suffix reuse mode. 
Using experience templates makes it possible to recognize that suffix actions can also be merged into the same state (e.g., an invariant step), 
thereby enabling suffix reuse.
For each experience template, the Action Memory module maintains an ActChain structure that stores detailed action sequences corresponding to the specific tasks mapped to that template. 
When a new task executes invariant steps, the Action Memory module directly reuses the action. 
For variable steps, if the associated parameter values match those of a historical task, 
the corresponding action can also be reused; otherwise, the LLM performs fresh reasoning and action generation.}

\textit{Correctness check and rollback mechanism.}
\FEH{To guarantee that cached actions remain valid despite potential app updates or UI changes, 
we employ a verification mechanism for detecting stale action memory before execution.
The Action Memory module examines the UI hierarchy (e.g., XML) to locate an element with matching properties such as resource ID, class name, and text content, 
and uses fuzzy matching to tolerate minor variations.}
When the check fails, we infer that the page layout has likely changed due to app updates or dynamic content, 
triggering a rollback mechanism that discards the failed action, falls back to the LLM execution and updates the Action Memory with the new execution trace.

	\nopagebreak
	\section{Agent Service: System Integration}
\label{sec:integration}

\begin{figure}[t]
  \centering
  \setlength{\abovecaptionskip}{-1pt}
  \setlength{\belowcaptionskip}{-15pt}
  \includegraphics[width=\columnwidth]{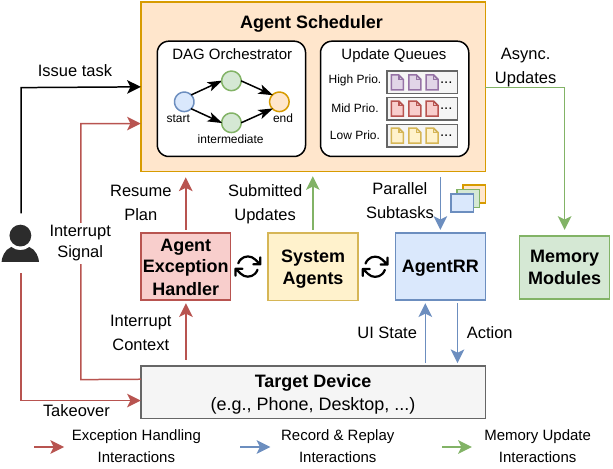}
  \caption{Agent Service integration architecture, showing how the Agent Scheduler, AgentRR, and Agent Exception Handler coordinate with memory modules, 
  system agents, and the target device.}
  \label{fig:integration}
\end{figure}

\FEH{By leveraging three types of agent memory, 
\systemname enables improvements in personalization, capability and execution efficiency during agent deployment without requiring additional model training. 
As shown in Figure~\ref{fig:integration}, inspired by task and memory management mechanisms in traditional operating systems, 
\systemname further incorporates agent-oriented scheduling, record-and-replay, interruption handling and exception recovery mechanisms. 
These enhancements improve the agent system's efficiency and robustness when handling multitasking scenarios and abnormal cases.}

\subsection{Agent Scheduler}
\label{subsec:scheduler}

\FEH{During task execution, both agent and memory operations incur latency, 
and a strictly serial workflow yields higher end-to-end latency. 
For example, memory retrievals can be completed within tens of milliseconds using embedding models and graph traversal, 
whereas memory updates require LLM inference, taking 1--2 seconds for profile construction and several seconds for template distillation. 
For operator agents, each task requires reading information from the screen, which takes about 1--10 seconds on average, 
depending on hardware capabilities and model size. 
However, these operations are not strictly interdependent, creating opportunities for parallel execution. 
In particular, this parallelism can be exploited during the planning stage, the execution stage, as well as enabling frontend-backend concurrency.
}

\textit{Parallel execution coordination.}
\FEH{The scheduler coordinates parallel execution at multiple granularities during task processing.
In the planning phase, \systemname concurrently retrieves user preferences and experience templates from the profile memory and experience memory. 
It then invokes the Task Rewriter to generate concrete task descriptions.
As for the execution phase, one task may involve multiple applications, 
\systemname supports both coarse-grained and fine-grained parallelism.
Coarse-grained parallelism enables parallel execution at the application level. 
When sub-tasks from different applications are independent, 
such as performing price comparisons across multiple shopping apps,
\systemname schedules them to run concurrently.
Fine-grained parallelism enables concurrency at the step level. 
For example, experience templates may indicate that app B only depends on a particular step of app A 
where a specific parameter value is required. 
In such cases, \systemname constructs a step-level DAG to capture these dependencies 
and maximize the concurrency of independent steps.
}

\textit{Background update prioritization.}
\FEH{The scheduler manages continuous background updates 
through three separate queues with different priorities. 
Profile updates are assigned the lowest priority because 
they do not directly affect the accuracy or efficiency of subsequent task execution. 
In contrast, updates to the Experience Memory module and Action Memory module are given higher priority, 
as they are triggered upon task completion 
and have immediate impacts on the performance of future tasks. 
This prioritization ensures that high-priority updates are processed promptly 
and are not delayed by the continuous stream of profile updates.}

\subsection{AgentRR: Agent Record and Replay}
\label{subsec:agentrr}

\FEH{The Action Memory mechanism relies on execution traces to support action reuse, 
but naïve action logging is insufficient for agent replay. 
The Action Memory module employs two structures, ActTree and ActChain; 
however, the ActTree must be dynamically constructed during execution, 
while the ActChain requires explicit mappings between experience templates and action sequences. 
We design Agent Record and Replay (AgentRR) to capture UI states, decision contexts, and execution steps necessary for both replay structures. 
Unlike traditional system-level record-and-replay techniques that merely reproduce exact action sequences, 
AgentRR offers agent-level replay capabilities that not only regenerate action trajectories 
but also preserve an agent's generalization ability through template-based abstraction and adaptive action verification.}

\textit{Recording mechanism.} 
\FEH{We design AgentRR as a lightweight instrumentation layer that intercepts all interactions between the agent and the mobile device. 
At each decision point during execution, AgentRR records the current UI state (including the screenshot and UI hierarchy) and the corresponding action.
For the ActTree, each recording adds a new path. 
If the nodes (UI states) and edges (actions) on this path match existing nodes and edges in the ActTree, 
AgentRR merges them and updates the associated task list.
For the ActChain, AgentRR summarizes the action sequence into an experience template and generates the corresponding step-to-action mappings. 
If an experience template already exists, the current action trajectory is used to validate it.}

\textit{Replay mechanism.} 
When replaying cached actions, AgentRR verifies each action before execution as described in \S\ref{subsec:action-cache}.
Upon successful verification, AgentRR translates the cached action into concrete operations.
Each cached action already contains the action type, target UI element, and parameters, 
enabling AgentRR to map it directly to the corresponding UI element in the current UI hierarchy.
If verification fails, AgentRR falls back to the Operator Agent for re-planning, 
allowing the agent to complete the remaining steps.
\FEH{After task execution, AgentRR will record the newly generated actions as an alternative path or update stale action memory caused by UI changes or application updates.
}

\subsection{Agent Exception Handler}
\label{subsec:exception-handling}

During task execution, users may need to interrupt the agent when they observe incorrect actions or want to make manual corrections.
Without proper exception handling mechanisms, such interruptions would break the automated execution flow.
\FEH{Moreover, if exception handlers are not preserved, 
the same errors may recur in subsequent executions, making it impossible to leverage prior handling experience.}

\textit{Exception detection and execution suspension.}
We design an exception-aware execution model that treats user interventions as recoverable events.
The system continuously monitors for interruption signals during execution, 
including explicit pause commands and manual UI interactions that conflict with planned agent actions.
When an interruption is detected, the scheduler immediately suspends the current execution thread and retains the context of the agent's execution process.
This includes preserving the full execution context: the current UI state, the partially completed task plan, and the action history.
The system then yields control to the user, allowing them to inspect the current state, review planned actions, 
and make corrections through natural language instructions or direct UI manipulations.

\textit{Recovery and continuation.}
\FEH{Once the user completes the corrections,
the system integrates the exception handling actions with the execution context to proceed with subsequent planning and execution.
Rather than directly incorporating user corrections into experience templates, 
the \textit{Experience Generator} analyzes the entire execution trace including both the agent's actions and user's corrections.}
It compares the original plan with the corrections, identifies where and why the agent deviated from user intent, 
and distills these insights into improved templates.
This analysis-driven approach ensures that the system learns what corrections the user made as well as 
why the corrections were necessary, enabling robust template evolution that prevents similar issues in future tasks.
	\nopagebreak
	\section{Evaluation}
\label{sec:evaluation}

    \begin{table}[t]
        \setlength{\abovecaptionskip}{0pt}
        \setlength{\belowcaptionskip}{5pt}
        \caption{Profile Memory Module evaluation results (500 historical tasks, 30 test tasks).}
    \centering
    \small
    \begin{tabular}{lccc}
    \hline
    \textbf{System} & \textbf{Write (ms)} & \textbf{Retrieval (ms)} & \textbf{Alignment (\%)} \\
    \hline
    Vanilla RAG & 1.76 & 19.58 & 66.4 \\
    GraphRAG & 37688.73 & 6675.82 & 81.1 \\
    Ours (Graph) & 6138.87 & 23.83 & 83.1 \\
    \hline
        \end{tabular}\\[-5pt]
    \label{tab:user-profile-results}
    \end{table}

    \begin{table}[t]
        \setlength{\belowcaptionskip}{5pt}
        \setlength{\abovecaptionskip}{0pt}
        \caption{Scalability of Profile Memory Module across different node counts.}
        \centering
        \small
        \begin{tabular}{lcccc}
        \hline
        \textbf{Nodes} & \textbf{Update} & \multicolumn{2}{c}{\textbf{Retrieval (ms)}} & \textbf{Storage} \\
        \cmidrule(lr){3-4}
        & \textbf{(ms)} & Vector DB & DisGraph & \textbf{(MB)} \\
        \hline
        100 & 6630.60 & 8.77 & 0.15 & 1.41 \\
        1{,}000 & 7051.38 & 21.11 & 0.17 & 13.49 \\
        10{,}000 & 8017.17 & 178.40 & 0.26 & 134.79 \\
        100{,}000 & 9193.84 & 1291.28 & 0.35 & 1346.48 \\
        \hline
        \end{tabular}\\[-5pt]
        \label{tab:profile-scalability}
    \end{table}

    We evaluate \systemname along three dimensions that mirror our system design in Section~\ref{sec:design}: 
    (1) whether the DisGraph-based User Profile module can efficiently capture and serve user preferences, 
    (2) how experience templates in the Experience Memory module affect task success rates and cost, 
    and (3) how the Action Memory module improves end-to-end latency across different tasks.
    \FEH{We deploy \systemname on both edge and cloud environments and conduct evaluations in real-world mobile usage scenarios, 
    covering the 50 most commonly used applications.
    Due to the strict latency requirements in edge scenarios, 
    real deployments still rely on cloud compute resources (Intel Xeon Platinum 8378A CPUs and NVIDIA A100-SXM4-80GB GPUs) for model inference.
    For on-device evaluation, we use the Qualcomm Snapdragon 8 Elite SoC platform and deploy the model using llama.cpp~\cite{llamacpp}.}
    We evaluate multiple SOTA GUI agent models for the operator agent to ensure generality, 
    including MobiMind-4B~\cite{zhang2025mobiagent}, UI-TARS-1.5-7B~\cite{qin2025uitars}, 
    GUI-Owl-7B~\cite{ye2025mobileagent}, Qwen3-VL-30B-A3B~\cite{bai2025qwen3vl} and Gemini-2.5-Flash~\cite{gemini25flash}, 
    covering different scales and architectures. 
    We do not evaluate industrial flagship reasoning models such as Gemini-2.5-Pro or open-source models with more than 10B activated parameters, 
    as their high inference latency cannot meet the latency SLOs of on-device agents.
    Other agents, such as Profile Updater, Experience Generator, and Task Rewriter, 
    are not on the critical path, and we adopt Qwen3-VL-30B-A3B as the base model.
    \FEH{The Experience Memory and AgentRR features proposed in \systemname have already been deployed on a flagship smartphone.}

    \begin{figure}[t]
        \centering
        \setlength{\abovecaptionskip}{-1pt}
        \setlength{\belowcaptionskip}{-10pt}
        \includegraphics[width=\columnwidth]{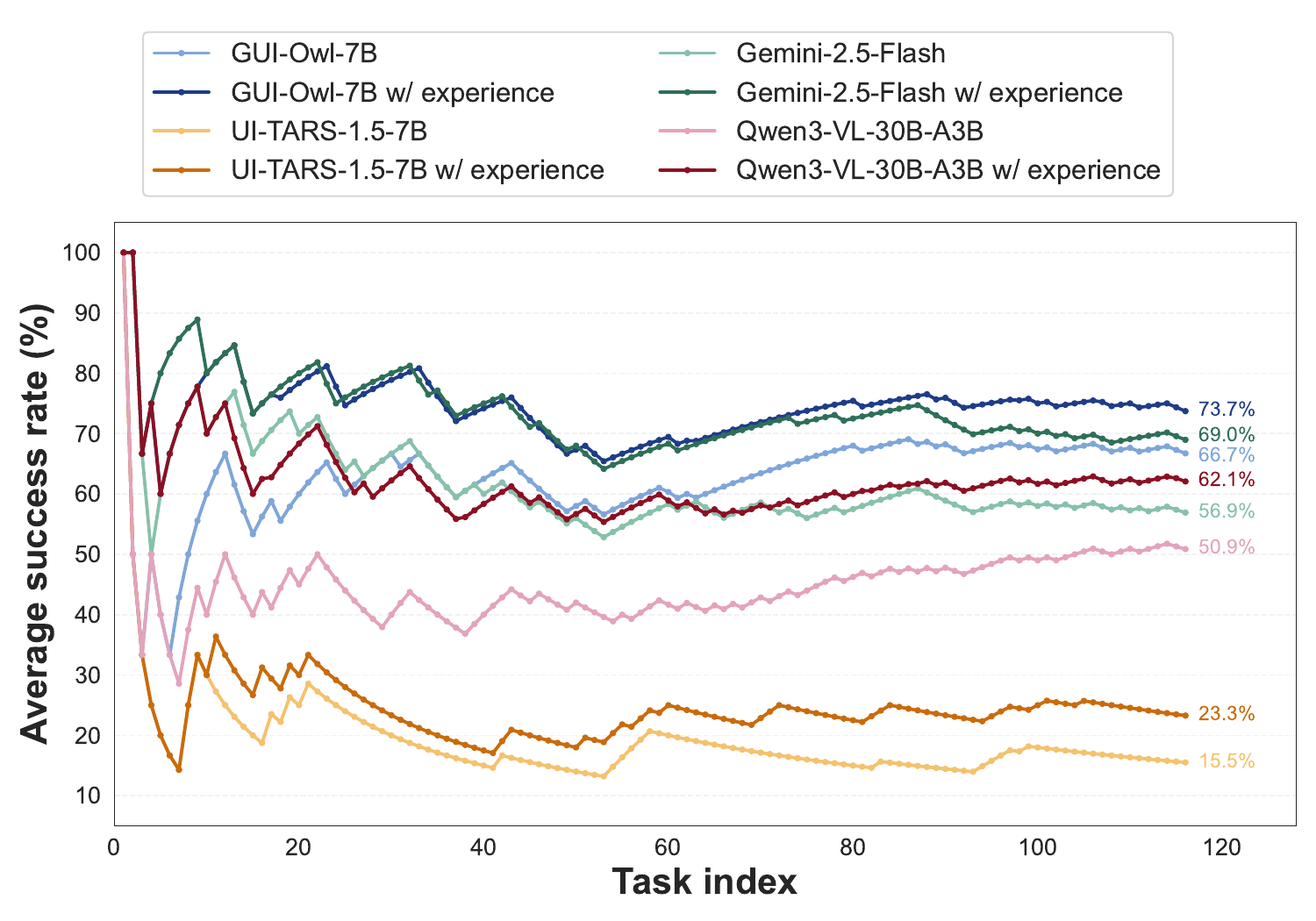}
        \caption{Average success rates of different agents on AndroidWorld with and without experience templates.}
        \label{fig:experience-template-accuracy}
    \end{figure}

    \begin{table}[htp]
        \setlength{\abovecaptionskip}{0pt}
        \setlength{\belowcaptionskip}{5pt}
        \caption{Efforts required to enable new agent capabilities when employing different approaches.}
        \centering
        \small
        \begin{tabular}{lcccc}
        \hline
        \textbf{Approach} & \textbf{Data} & \textbf{Person-hrs} & \textbf{GPU hrs} & \textbf{Acc.} \\
        \hline
        Fine-tuning & $\sim$100 ex. & 4.0 & 0.25 & 58.5\% \\
        Exp.(Man.) & $\sim$\textbf{5 ex.} & 0.2 & \textbf{0} & \textbf{63.5\%} \\
        Exp.(Syn.) & $\sim$\textbf{5 ex.} & \textbf{0} & 0.0027 & 60.1\% \\
        \hline
        \end{tabular}\\[-5pt]
        \label{tab:task-support-cost}
    \end{table}

    \subsection{Agent's Personalization}
    \label{subsec:eval-user-profile}

    \begin{figure}[t]
        \centering
        \setlength{\abovecaptionskip}{-1pt}
        \setlength{\belowcaptionskip}{-10pt}
        \includegraphics[width=\columnwidth]{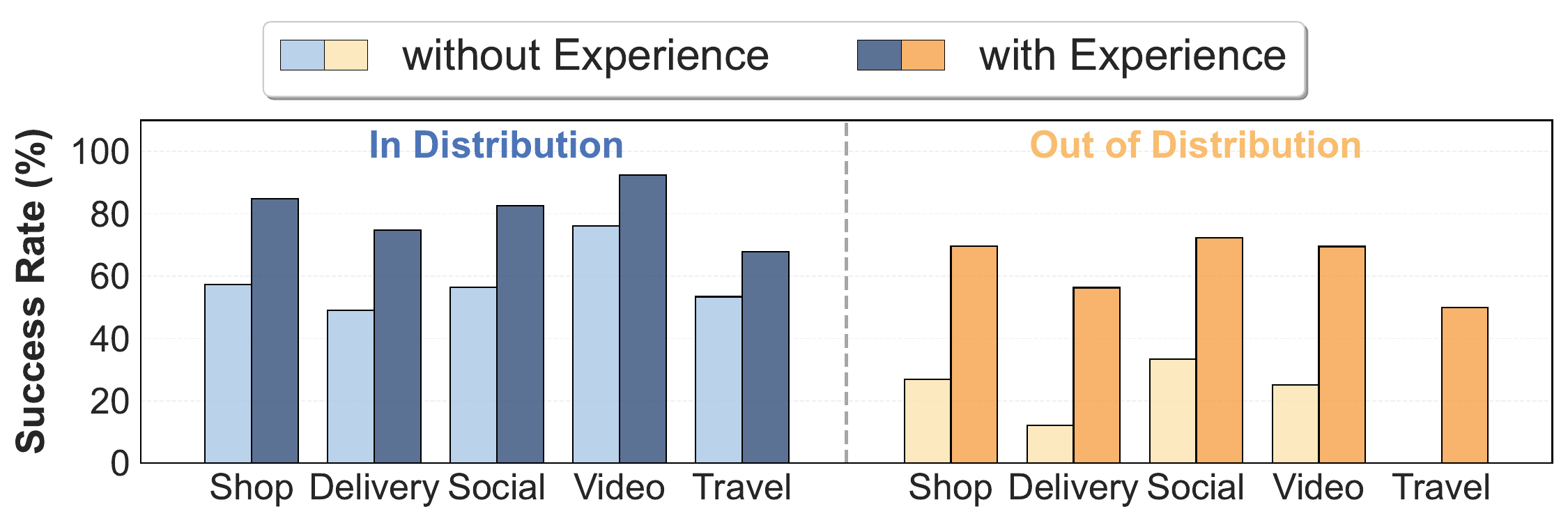}
        \caption{Task success rates with and without experience templates in ID/OOD scenarios.}
        \label{fig:experience-boost}
    \end{figure}

    \begin{figure}[t]
        \centering
        \setlength{\abovecaptionskip}{-1pt}
        \setlength{\belowcaptionskip}{-15pt}
        \includegraphics[width=\columnwidth]{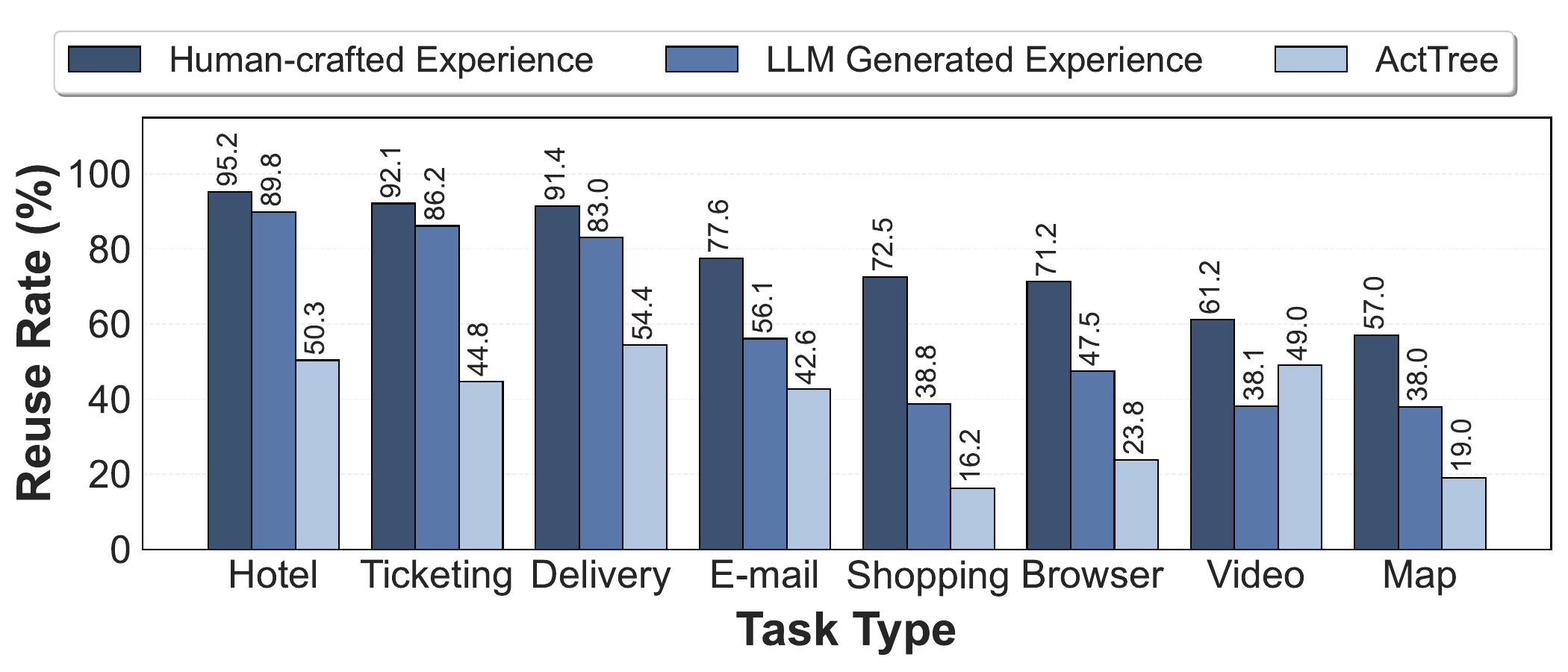}
        \caption{Action reuse rates of ActTree and ActChain with LLM-generated/human-crafted experience templates.}
        \label{fig:replay-rates}
    \end{figure}
    To evaluate whether agents can accurately learn and model user profiles from historical task executions, 
    we design a benchmark\footnote{Details in the Appendix.} with synthetic user profiles.
    Each profile contains descriptions of a person's characteristics across various aspects, 
    including facts (e.g., home address, frequently used accounts), 
    preferences (e.g., shopping habits, hotel booking preferences), etc.
    For each user, we generate 500 historical tasks that explicitly reflect the user's profile information, 
    and 30 test tasks with ambiguous descriptions that require the agent to infer missing details based on learned profile memory.
    For each test task, we define which profile information should be retrieved based on the task context, 
    serving as the ground truth for evaluating retrieval quality.
    During evaluation, each system learns from the historical tasks by updating its profile store, 
    then retrieves relevant profile information for the test tasks.
    We measure three metrics: write latency (time to update profile), 
    retrieval latency (time to retrieve profile information), 
    and profile alignment (the ratio of the retrieved profile information to the ground truth).
    We compare our DisGraph-based Profile Memory against two baselines: \textit{Vanilla RAG}, 
    which stores preferences as unstructured text in a vector database with embedding-based retrieval, 
    and \textit{GraphRAG}, which constructs a knowledge graph with entity extraction and hierarchical summarization following the Mem0 approach.
    All approaches use a context window of 2000 tokens for retrieval.

    As shown in Table~\ref{tab:user-profile-results}, Vanilla RAG achieves low latency 
    (1.76ms write, 19.58ms retrieval) but suffers from poor alignment (66.4\%) due to its flat structure lacking relational context.
    GraphRAG improves alignment to 81.1\% by encoding relational semantics in edge weights, 
    but incurs high latency (37.69s write, 6.68s retrieval) from expensive LLM calls for entity extraction, edge management, and query traversal.
    Our system achieves the best of both compared to the GraphRAG: 83.1\% alignment (25\% higher than Vanilla RAG) 
    with 6.14s write and 23.83ms retrieval latency (6.1$\times$ and 280$\times$ faster than GraphRAG respectively).
    \FEH{This stems from our DisGraph architecture that decouples relational semantics from edges to nodes: 
    writes require only one LLM call for entity classification and concept attachment, 
    while retrieval operates through embedding-based starting point selection and BFS traversal with zero LLM calls.
    High accuracy is maintained because multi-branch BFS gathers contextually relevant information from multiple conceptual dimensions.}

    \textbf{Scalability analysis.}
    To evaluate how Profile Memory scales with user profile size,
    we test update and retrieval performance as the number of profile nodes grows from 100 to 100,000.
    As shown in Table~\ref{tab:profile-scalability},
    update latency increases moderately from 6.6 seconds to 9.2 seconds as profile size grows.
    \FEH{Because the update process is dominated by a single LLM inference for entity classification and concept attachment, 
    its runtime remains nearly unchanged.}
    For retrieval, DisGraph traversal remains nearly constant at about 0.15ms--0.35ms across all profile sizes,
    confirming the zero-LLM overhead of our graph walking mechanism.
    \FEH{Total retrieval latency is dominated by the initial vector database search used to select starting nodes, 
    which increases from 8.8 ms (100 nodes) to 1.29 s (100,000 nodes, with 135 MB storage), a delay that remains acceptable in real-world scenarios.}

    \subsection{Agent's Capability}
    \label{subsec:eval-experience-template}

    \textbf{Task success rate in AndroidWorld.} 
    We first evaluate how experience templates improve task success rates in AndroidWorld~\cite{rawles2024androidworld}, 
    which provides 116 diverse tasks across 20 Android apps.
    Our baseline implementation references DroidRun~\cite{droidrun}, the SOTA open-source work on AndroidWorld, 
    and we augment it with our experience mechanisms to measure the improvements.
    We test four agent models with and without experience templates: GUI-Owl-7B, 
    UI-TARS-1.5-7B\footnote{The 33.0\% accuracy is reported in UI-TARS paper, but we cannot fully reproduce this result as the prompts used for testing are not open-sourced.}, 
    Gemini-2.5-Flash, and Qwen3-VL-30B-A3B.

    \FEH{As shown in Figure~\ref{fig:experience-template-accuracy}, 
    experience templates consistently improve success rates across all agents. 
    Weaker models benefit more: UI-TARS-1.5-7B achieves a 50.3\% relative improvement, 
    while stronger models such as Gemini and Qwen gain 21\%--22\%, and GUI-Owl gains 10.5\%. 
    Beyond this overall trend, we observe that different template levels contribute differently across model types. 
    For general-purpose models such as Gemini and Qwen, lower-level templates are more effective 
    because the logic of UI interactions is difficult to infer through reasoning alone.
    In contrast, for domain-specific models like GUI-Owl, which already possess basic UI interaction skills, 
    higher-level templates reduce reliance on autonomous planning, thereby yielding greater benefits for smaller models.}
    Each experience template consumes 7.8KB on average, resulting in 900KB total storage for the 116 templates.

    \begin{figure}[t]
        \centering
        \setlength{\abovecaptionskip}{-1pt}
        \setlength{\belowcaptionskip}{-15pt}
        \includegraphics[width=\columnwidth]{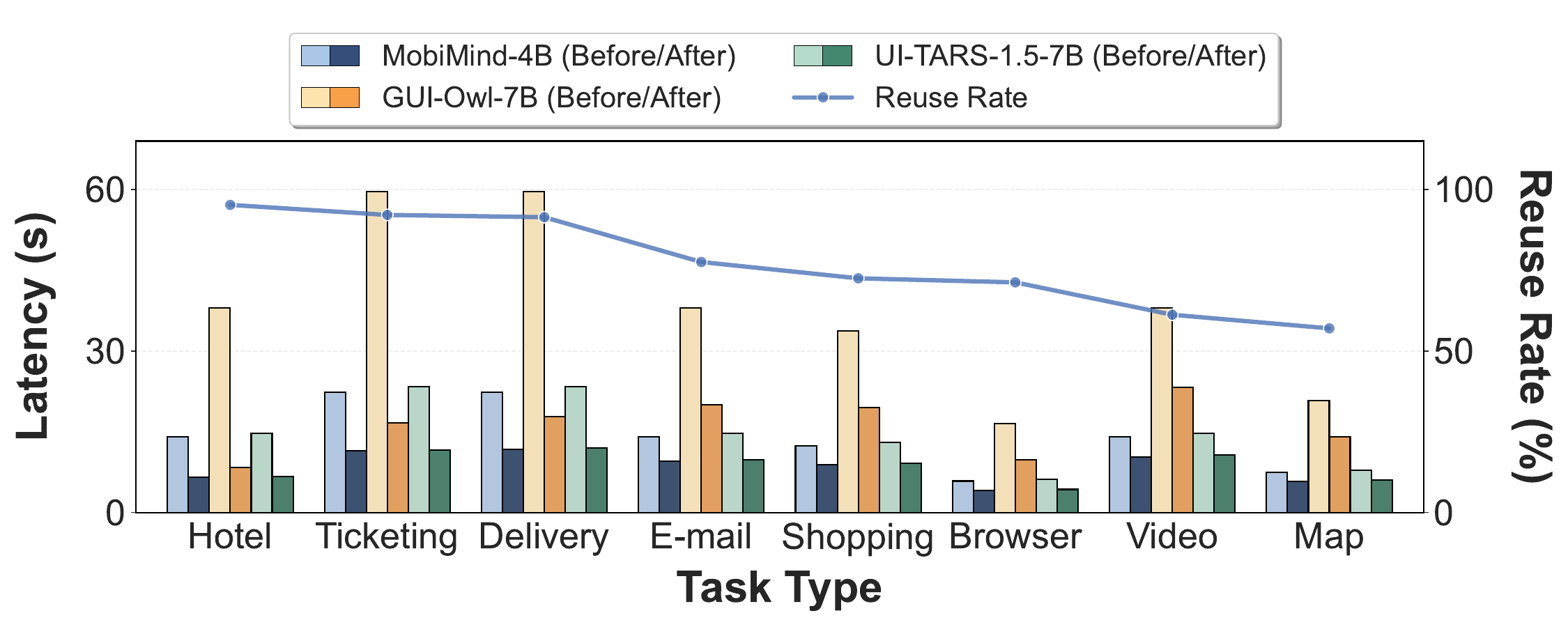}
        \caption{End-to-end latency and reuse rates for three agent models with and without Action Memory.}
        \label{fig:experience-rr}
    \end{figure}

    \textbf{Task success rate in real-world applications.}
    We evaluate success rates when applying experience memory across various real-world tasks, 
    covering both in-distribution (ID) and out-of-distribution (OOD; i.e., long-tail tasks beyond the agent's training data). 
    UI-TARS-1.5-7B is used as the agent model for this evaluation. 
    For tasks that are only partially completed, the completion rate is taken as the success rate. 
    \FEH{As shown in Figure \ref{fig:experience-boost}, 
    applying experience memory leads to substantial improvements in task success rates for both OOD and ID settings. 
    Because OOD tasks often involve complex reasoning, 
    introducing experience templates significantly reduces the reasoning burden and yields greater accuracy gains 
    (44.1\% for OOD tasks and 22.0\% for ID tasks).}

    \textbf{Cost-effectiveness.}
    To evaluate the cost-effectiveness of experience memory, we compare three approaches for enabling new agent capabilities: model fine-tuning, manually authored experience templates, and automatically synthesized experience templates in our system.
    Table~\ref{tab:task-support-cost} compares these approaches across four key metrics: data volume (number of task execution traces needed), person-hours (human effort for data collection, annotation, or template authoring), GPU hours (computational resources for model training or LLM-based template generation), and accuracy (task success rate).

    Fine-tuning requires $\sim$100 training examples, 4 person-hours, and 0.25 GPU hours to achieve 58.5\% accuracy, but must be repeated for each new task family.
    In contrast, both template methods require only $\sim$5 examples (20$\times$ less data) by explicitly separating invariant control logic from variable parameter slots.
    Manual authoring takes 0.2 person-hours for experts to identify control logic and mark variable slots, achieving the highest accuracy (63.5\%) through more precise separation of invariant and variant steps, while automatic synthesis automates this process through semantic clustering and action sequence alignment, requiring zero human effort and only 0.0027 GPU hours to achieve 60.1\% accuracy.
    This demonstrates that experience memory provides a cost-effective alternative to fine-tuning by separating invariant control logic from runtime variables, enabling rapid task family coverage while maintaining competitive accuracy.

    \subsection{Agent's Efficiency}
    \label{subsec:eval-action-memory}

    To evaluate the effectiveness of action reuse, \FEH{we evaluate the different structures of Action Memory under realistic workload patterns,
    which include 8 diverse task categories}: email, train ticketing, food delivery, hotel booking, shopping, web browser, media playback, and map navigation.
    For each category, we generate parameterized task instances using an LLM to create realistic variations.
    We submit a sequence of 454 tasks to the system across these categories.

    \begin{figure}[t]
        \centering
        \setlength{\abovecaptionskip}{-1pt}
        \setlength{\belowcaptionskip}{-15pt}
        \includegraphics[width=\columnwidth]{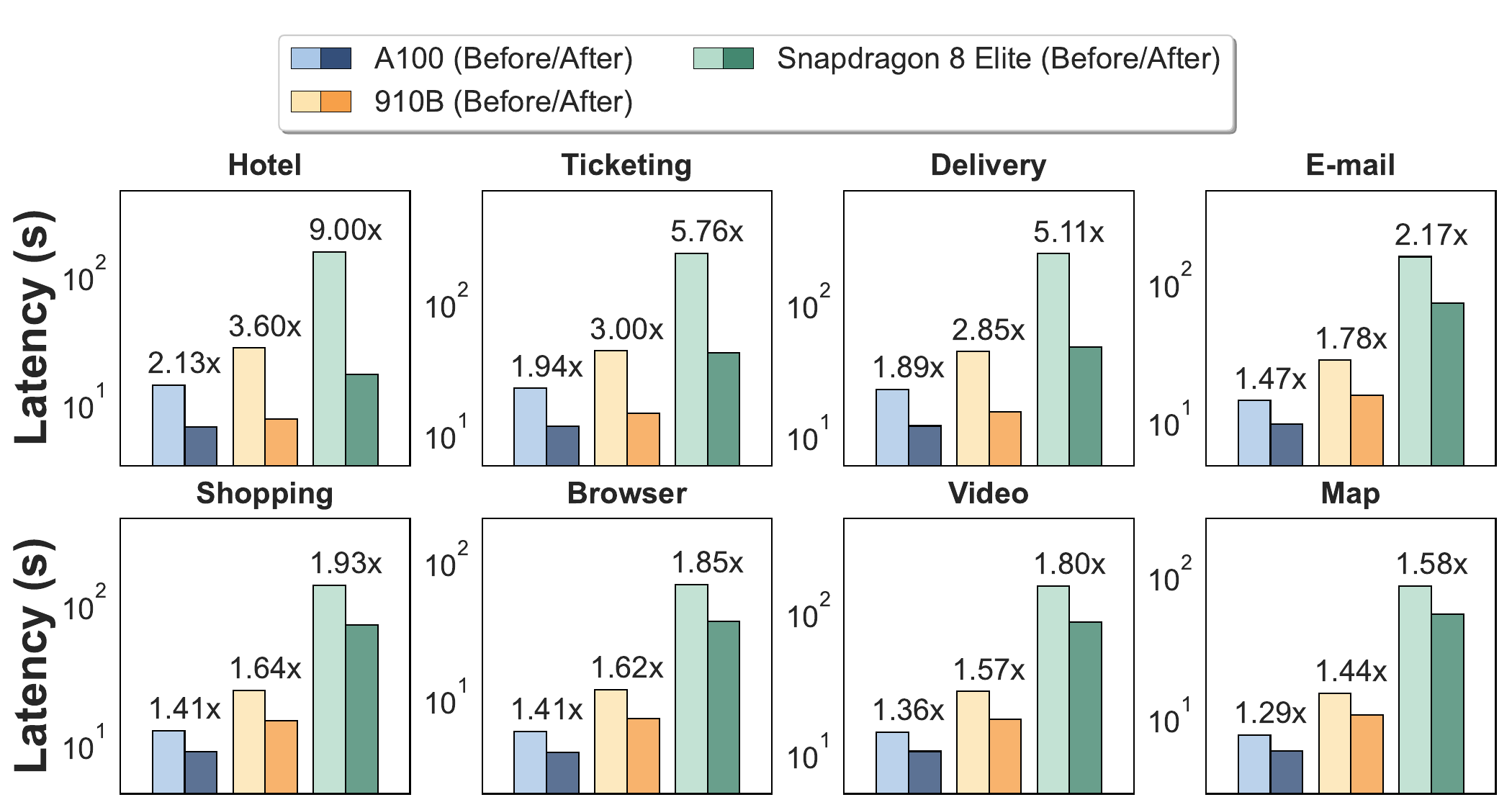}
        \caption{Performance boost of Action Memory across different hardware configurations.}
        \label{fig:hardware-comparison}
    \end{figure}

    \textbf{Action reuse rate.} 
    We first evaluate how different implementations of Action Memory affect action reuse rates.
    Figure~\ref{fig:replay-rates} compares the reuse rates of ActTree and ActChain across task categories.
    For ActChain, we further evaluate the impact of two template sources on reuse rates: LLM-generated templates and human-crafted templates.
    ActTree only exploits prefix reuse, achieving 37.5\% average reuse rate.
    ActChain exploits both prefix and suffix reuse, reaching higher reuse rates compared to ActTree.
    Human-crafted templates strictly separate invariant and variable actions into adjacent steps to maximize the reuse of invariant steps, 
    achieving the highest average reuse rate of 77.3\%.
    LLM-generated templates do not enforce the above constraint, 
    resulting in lower average reuse rate of 59.7\% but still outperform ActTree, demonstrating the effectiveness of prefix-suffix reuse.

    \textbf{End-to-end latency.} 
    We evaluate how Action Memory affects end-to-end task execution time across different agents.
    We employ ActChain with human-crafted templates for this evaluation.
    As shown in Figure~\ref{fig:experience-rr}, without Action Memory, latency varies across agents due to differences in inference efficiency: 
    MobiMind-4B averages 14.1s, UI-TARS-1.5-7B 14.7s, and GUI-Owl-7B 38.0s.
    With Action Memory enabled, all three models achieve substantial latency reductions: MobiMind-4B reduces to 8.6s (up to 2.1$\times$ speedup), 
    UI-TARS-1.5-7B to 8.8s (up to 2.2$\times$ speedup), and GUI-Owl-7B to 16.2s (up to 4.5$\times$ speedup).
    The latency reduction correlates strongly with reuse rates: tasks such as hotel query and train ticketing with over 92\% reuse rates achieve up to 4.5$\times$ speedup, 
    demonstrating that Action Memory effectively eliminates the bottleneck of LLM inference.
    Tasks such as shopping and browser with lower reuse rates ($\sim$70\%) still require longer execution time due to more frequent LLM inference calls for cache misses. 
    For memory overhead, Action Memory uses only 1.54MB ($\sim$6,000 cached actions), negligible on modern devices.

    \begin{figure}[t]
        \centering
        \setlength{\abovecaptionskip}{-1pt}
        \setlength{\belowcaptionskip}{-15pt}
        \includegraphics[width=\columnwidth]{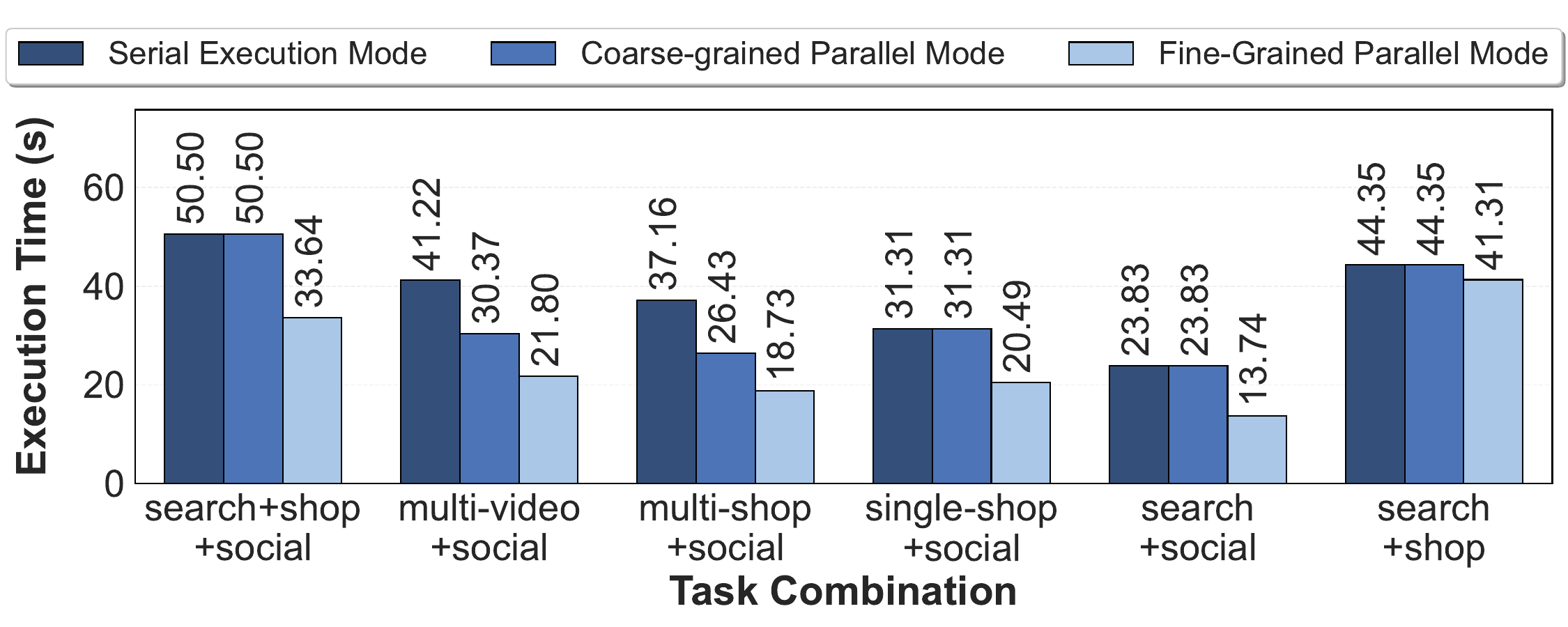}
        \caption{End-to-end performance of different execution modes across six task categories.}
        \label{fig:parallel-comparison}
    \end{figure}
    
    \textbf{Performance across different hardware.} 
    Finally, we evaluate Action Memory across different hardware configurations representing diverse deployment scenarios: A100 GPU, Ascend 910B NPU~\cite{ascend} (cloud server), 
    and Qualcomm Snapdragon 8 Elite SoC~\cite{snapdragon8elite} (mobile device).
    We use MobiMind-4B as the agent model for this evaluation.
    As shown in Figure~\ref{fig:hardware-comparison}, without Action Memory, task latency varies dramatically across hardware: 
    A100 averages 14.1s, 910B 27.4s, while Snapdragon suffers from 153.2s due to limited computational resources (CPU-only).
    After enabling Action Memory, latency becomes much more uniform across hardware configurations: 
    most tasks complete within 10s--50s across different underlying compute platforms.
    The speedup is most pronounced on mobile device where Action Memory achieves 1.6$\times$--9$\times$ speedup across tasks, 
    as action reuse effectively eliminates expensive on-device inference.
    Even on high-performance hardware like A100, Action Memory provides consistent 1.3$\times$--2.1$\times$ speedup by avoiding redundant inference.
    This demonstrates that Action Memory effectively shifts the bottleneck from LLM inference to lightweight action execution and UI interaction, 
    enabling practical deployment on mobile and edge devices.

    \begin{figure}[t]
        \centering
        \setlength{\abovecaptionskip}{-1pt}
        \setlength{\belowcaptionskip}{-15pt}
        \includegraphics[width=\columnwidth]{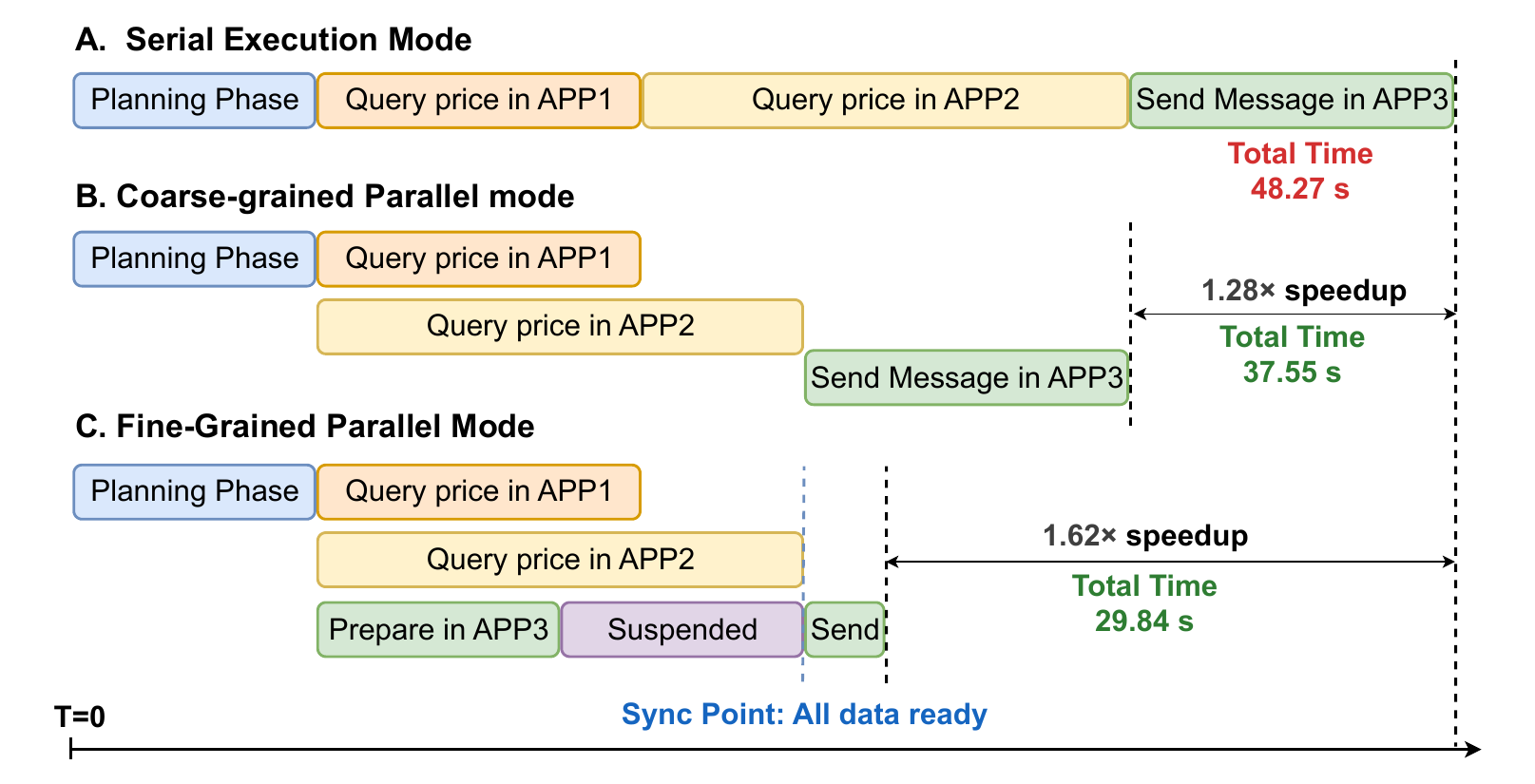}
        \caption{Detailed execution timeline comparison between serial and parallel modes for a multi-shop+social task.}
        \label{fig:parallel-execution-analysis}
    \end{figure}
    
    \subsection{Performance on Multi-task Scheduling}
    \label{subsec:eval-end-to-end}

    To evaluate \systemname's performance in realistic scenarios, we test \systemname across complex multi-app tasks 
    that require cross-app coordination and data transfer.
    These tasks represent real-world use cases where information is retrieved from one or more apps and consumed in other apps.
    The benchmark involves three application categories: search, shop, and social networking, using applications such as Chrome, Amazon, WeChat, etc.
    These application categories are combined to form six task categories as shown in Figure~\ref{fig:parallel-comparison}, 
    where the ``multi-'' prefix indicates that multiple apps of the same type are involved.
    Each category contains multiple task instances with different parameter initializations.\footnote{Details in the Appendix.}
    We compare the end-to-end latency of three execution modes: 
    \textit{Serial} executes all sub-tasks sequentially.
    \textit{Coarse-grained} parallelizes independent sub-tasks at the application level.
    \textit{Fine-grained} further enables step-level parallelism within each sub-task.

    Coarse-grained parallelism achieves up to 1.41$\times$ speedup over serial execution in the multi-shop+social task.
    Fine-grained parallelism achieves more substantial improvements, with up to 1.98$\times$ speedup over serial execution in the same task category.
    For tasks with sequential app dependencies (e.g., search+shop+social), 
    coarse-grained parallelism provides no benefit, while fine-grained mode still achieves 1.50$\times$ and 1.53$\times$ speedups through step-level parallelism.

    To understand the performance improvements, we analyze a representative task from the multi-shop+social category, 
    as illustrated in Figure~\ref{fig:parallel-execution-analysis}.
    This task queries prices for the same item across two shopping apps and then shares the results via a social app.
    In serial mode, the system sequentially queries prices in two shopping apps and then sends a message in a social app, taking 48.27 seconds total.
    In coarse-grained mode, the two price queries execute in parallel, reducing execution time to 37.65 seconds (1.28$\times$ speedup).
    In fine-grained mode, the scheduler proactively executes steps in the social app that do not depend on price data (e.g., navigating to the chatting page) 
    while price queries run in parallel, blocking only at message composition when price data is required.
    This further reduces execution time to 29.84 seconds (1.62$\times$ speedup), 
    demonstrating that \systemname's fine-grained parallelization effectively exploits step-level dependencies and achieves superior performance.

	\nopagebreak
	\section{Discussion}

\textbf{Platform portability.}
Our system is implemented on mobile devices with structured GUI introspection, OS-level execution APIs, and execution trace logging.
The evaluation demonstrates hardware portability across cloud GPUs, NPUs, and mobile SoCs, validating that our memory mechanisms maintain effectiveness across diverse computational resources.
The core memory mechanisms (DisGraph for user profiles, multi-level templates for experiences, and ActTree/ActChain for action reuse) are platform-agnostic abstractions that can be adapted to different environments.
Desktop systems provide equivalent capabilities through Microsoft UI Automation~\cite{windows}, Linux AT-SPI~\cite{atspi}, and Apple Accessibility API~\cite{macos}, web browsers offer comprehensive DOM introspection with tools like Selenium~\cite{selenium} and Playwright~\cite{playwright}, and command-line interfaces can leverage process monitoring and output parsing.
Through appropriate abstraction layers, our memory mechanisms can generalize across diverse platforms.

\textbf{Comparison with general agent systems.}
General agent frameworks such as LangChain~\cite{chase2022langchain}, AutoGen~\cite{wu2024autogen}, and MetaGPT~\cite{hong2023metagpt} orchestrate multi-agent workflows atop LLM serving systems~\cite{zhong2024distserve,sun2024llumnix,kwon2023vllm,lin2024parrot,yu2022orca,fu2024serverlessllm,zheng2022alpa,sarathi-serve,zheng2024sglang} that optimize inference efficiency.
While these frameworks enable flexible agent composition, their memory mechanisms target conversational agents with linear interaction patterns rather than GUI agents operating in stateful visual environments.
The higher cost and complexity of GUI automation necessitate specialized memory: costly physical UI operations require fine-grained action caching, trial-and-error in long-horizon tasks demands abstraction of reusable experience templates from execution traces, and personalization requires capturing multi-app user preference patterns.
Existing frameworks rely on conversation buffers designed for dialogue history rather than memory supporting state-indexed action retrieval, hierarchical experience abstraction, and evolving user profiles.
\systemname introduces three specialized memory types that transform GUI agents from static executors into self-evolving systems that improve through deployment.

	\nopagebreak
	\section{Conclusion}

In this paper, we present \systemname, a memory-centric system that enables agent self-evolution without continual model training.
\systemname introduces three specialized memory types with system-level integration, demonstrating significant improvements in preference alignment, execution accuracy and latency.
As AI agents are increasingly deployed in edge environments, we believe the memory-centric paradigm offers a practical path toward continually evolving agents that learn from agent memories rather than expensive model updates.
\bibliographystyle{plain}
\bibliography{reference}
	
	\appendix
	\clearpage
	\setcounter{section}{0}
	\renewcommand{\thesection}{\Alph{section}}
	\renewcommand{\thesubsection}{\thesection.\arabic{subsection}}
	\renewcommand{\thesubsubsection}{\thesubsection.\arabic{subsubsection}}
	\section{User Profile Benchmark}
\label{sec:appendix-profile}

We design a user profile benchmark to evaluate whether agents can accurately learn and model user profiles from historical task execution traces.
The benchmark consists of 20 synthetic user profiles, each containing textual descriptions across various aspects of user behavior including facts (e.g., home address, work location), preferences (e.g., shopping habits, hotel booking preferences), and past experiences (e.g., previously visited places, completed activities).
For each user, we generate 500 historical tasks that explicitly reflect the user's profile information, and 30 test tasks with ambiguous descriptions that require agents to infer missing details based on learned profile memory.
User profiles span 5 task categories (shopping, hotel booking, travel, food delivery, entertainment) with over 30 profile dimensions in total.
This section provides detailed information about benchmark design, task formats, LLM-based task rewriting, and evaluation methodology.

\subsection{Benchmark Design}

\subsubsection{Design Rationale}
The benchmark evaluates three core capabilities of agent systems.
First, agents must extract structured profile information from task execution traces containing implicit signals about user facts, preferences, habits, and experiences.
Second, agents must store this information in memory systems that enable efficient retrieval.
Third, agents must apply learned profiles to complete ambiguous tasks by inferring missing task parameters.

\subsubsection{Data Structure}
Each user profile contains structured information across five task categories: shopping, hotel booking, travel, food delivery, and entertainment.
Within each category, profiles specify 3--7 dimensions covering various aspects such as preferences, habits, constraints, and contextual information.
Historical tasks explicitly encode user profile information through natural language instructions.
Test tasks are intentionally ambiguous, omitting specific profile details to require agents to retrieve and apply learned profile memory.

\subsection{Task Format and Examples}

\subsubsection{Historical Task Format}
Historical tasks are natural language instructions that explicitly contain user profile information.
Each task is a complete sentence describing a specific action, with profile information (preferences, habits, constraints) embedded naturally.

\textbf{Example historical tasks for a budget-conscious user:}

\begin{tcolorbox}[colback=gray!5, colframe=gray!40, boxrule=0.5pt]
\small
\noindent
``I need to buy a keychain organizer on Taobao---looking for something affordable from a domestic brand, beige color preferred, under 20 yuan, and I'd like next-day delivery if possible.''

\medskip
\noindent
``Can you help me find high-speed rail tickets from Chengdu to Luzhou on the 12306 app? I'm visiting relatives and prefer a direct train around 11 AM, second-class seat is fine to keep costs down. I usually book three days ahead.''

\medskip
\noindent
``I'd like to order a light lunch on Eleme---Chinese food, but not too spicy. Something budget-friendly with beef brisket and lots of vegetables would be perfect. Please have it delivered during my lunch break.''
\end{tcolorbox}

\noindent
These tasks contain explicit profile signals including price preferences (``affordable'', ``under 20 yuan'', ``keep costs down'', ``budget-friendly''), brand preferences (``domestic brand''), style preferences (``beige color'', ``not too spicy''), timing constraints (``visiting relatives'', ``lunch break''), and behavioral patterns (``I usually book three days ahead'').

\subsubsection{Shared Test Task Format}
Shared test tasks are intentionally ambiguous, containing only high-level task descriptions without specific profile details.
Agents must infer and apply learned profile information to complete these tasks.

\textbf{Example shared test tasks:}

\begin{tcolorbox}[colback=gray!5, colframe=gray!40, boxrule=0.5pt]
\small
\noindent
``I'm heading to Guangzhou tomorrow to meet some friends. Can you book me a train ticket on 12306, find a hotel on Ctrip near good transit options for one night, and order some household items on Taobao that I can bring with me?''

\medskip
\noindent
``Planning to relax at home this weekend. Could you order some Chinese food for me on Meituan, find some interesting short videos on Bilibili, and maybe get me a comfy cushion from Taobao?''
\end{tcolorbox}

These tasks are ambiguous in several ways.
Price constraints are unspecified: ``book me a train ticket'' does not indicate seat class or budget.
Brand and style information is missing: ``order some household items'' does not specify brand, quality level, or style.
Timing and context are unclear: ``relax at home this weekend'' does not provide exact delivery windows, viewing schedules, or personal taste information.

\subsection{LLM-Based Task Rewriting}

To improve the realism of historical tasks, we employ an LLM-based rewriting process that transforms template-generated tasks into natural user utterances.
The rewriting process takes structured task templates and produces conversational language while preserving all profile signals embedded in the original tasks.

\subsubsection{Rewriting Process}
The rewriting process takes original tasks generated from templates along with user profile summaries as input.
The LLM is then prompted to rewrite these tasks into more natural language while preserving the core task intent and all embedded profile information.
The process produces rewritten tasks that maintain all profile signals but express them through conversational language that resembles authentic user requests.

\textbf{Example rewriting:}

\begin{tcolorbox}[colback=gray!5, colframe=gray!40, boxrule=0.5pt]
\small
\noindent
\textbf{Original}: ``Search for a pillow on Taobao, choose domestic brand and lowest price''.

\medskip
\noindent
\textbf{Rewritten}: ``I'm looking for an affordable pillow on Taobao---preferably memory foam from a domestic brand. Something really comfortable in beige would be great, and my budget is around 100 yuan. If you can get next-day delivery, that'd be perfect!''
\end{tcolorbox}

\noindent
The rewritten version maintains core profile signals (price, brand, style, delivery, timing), uses more natural and conversational language, adds contextual details that make the request sound authentic, and preserves all required profile dimensions while sounding like a real user query.

\subsection{Evaluation Methodology}

\subsubsection{Evaluation Workflow}
The evaluation follows a three-phase workflow.

\textbf{Phase 1: Learning}.
The agent processes historical tasks (500 per user), extracts profile information and stores it in memory, and learning is evaluated by checking if extracted profile information matches the ground-truth profile.

\textbf{Phase 2: Task Rewriting}.
The agent receives shared test tasks (30 ambiguous tasks), retrieves relevant learned profile information for each task, uses LLM to rewrite the ambiguous task into a personalized complete task description, and the rewritten task should incorporate learned profile information naturally.

\textbf{Phase 3: Profile Matching}.
For each rewritten task, an LLM judge evaluates whether required profile information is present, compares the rewritten task against the ground-truth profile, and computes metrics including matched information count, total information count, and alignment score.

\subsubsection{Required Profile Information Generation}

To determine what profile information should be present in each test task, we employ an LLM-based analysis process.
For each user, the process takes three inputs: the user's ground-truth profile, the user's historical task history, and the shared test tasks.
The LLM analyzes each test task and identifies which specific profile elements are necessary to complete that task for the given user.

\textbf{Example analysis prompt template:}

\begin{tcolorbox}[colback=gray!5, colframe=gray!40, boxrule=0.5pt]
\small
You are an expert in user profile analysis. For each task in the shared task list, determine what profile information is required to complete that task.

\textbf{User Profile:}

[Coarse-grained profile information across different categories]

\textbf{User Task History:}

[Sample historical tasks showing fine-grained behavior patterns]

\textbf{Shared Tasks:}

[Test tasks to analyze, numbered 1, 2, 3, ...]

\textbf{Output:}

For each task, identify the required profile information:

\begin{verbatim}
Task 1: { element_1: value, ... }
Task 2: { element_1: value, ... }
...
\end{verbatim}
\end{tcolorbox}

This analysis produces a structured mapping from each test task to its required profile information, represented as key-value pairs (e.g., ``price\_constraint: budget-friendly'', ``brand\_preference: domestic brand'').
This approach ensures that evaluation is context-aware: different test tasks may require different subsets of profile information, and the required information is determined based on both the task nature and the user's historical behavior patterns.

\subsubsection{Evaluation Metrics}

We evaluate systems using three metrics: write latency (time to update profile from each historical task), retrieval latency (time to retrieve profile information for each test task), and profile alignment (percentage of ground-truth profile information successfully retrieved, evaluated by an LLM judge).

For profile alignment evaluation, the system first identifies which profile information should be present in each test task based on task type and ground-truth profile.
An LLM judge then evaluates whether the agent's rewritten task contains these required profile elements.

\textbf{Example evaluation prompt template:}

\begin{tcolorbox}[colback=gray!5, colframe=gray!40, boxrule=0.5pt]
\small
You are an evaluator assessing whether a personalized task reflects required user profile information. Determine if the following profile elements are present in the rewritten task.

\textbf{Required Profile Information:}

[Profile information based on user profile and task context]

\textbf{Personalized Rewritten Task:}

[Rewritten task text]

\textbf{Instructions:}

Analyze the rewritten task and determine whether each profile element is clearly reflected.
\end{tcolorbox}

The judge returns a structured JSON response:

\begin{tcolorbox}[colback=gray!5, colframe=gray!40, boxrule=0.5pt]
\small
\begin{verbatim}
{
  "profile_check": [
    {
      "profile_element": "...",
      "expected_value": "...",
      "matched": true/false,
      "evidence": "..."
    },
    ...
  ]
}
\end{verbatim}
\end{tcolorbox}

\noindent
Each element includes the profile element name, expected value (from user profile), a boolean ``matched'' field indicating whether the profile information is reflected in the task, and an ``evidence'' field providing justification.

\textbf{Computed metrics}:
We compute per-task alignment as the percentage of required profile elements matched in each rewritten task, overall alignment as the average alignment across all 30 test tasks, and per-dimension accuracy as the matching accuracy for each profile dimension.
These metrics quantify both the completeness and accuracy of profile learning and retrieval.

	\nopagebreak
	\section{Multi-Task Execution in Real-World Scenarios}
\label{sec:appendix-multitask}

To complement the performance evaluation presented in our main paper, this appendix provides detailed descriptions of the real-world testing scenarios used to evaluate the Agent Scheduler's multi-task execution capabilities.
These scenarios involve complex workflows requiring data transfer and synchronization across multiple applications, representing typical user interactions in practical mobile usage.
We measure end-to-end execution latency across three execution modes: Serial, Coarse-grained Parallel, and Fine-grained Parallel.

\subsection{Execution Modes in Practice}

We use a representative shopping and social networking scenario to illustrate how the three execution modes handle real-world tasks differently.

\textbf{Scenario:} Query the price of a specific item in two different shopping applications (App A and App B), and then send the gathered information to a contact via a social networking application (App C).

\subsubsection{Serial Execution}
The agent executes tasks strictly sequentially.
It first completes the price query in App A, then performs the price query in App B, and finally launches App C to send the message.

\textit{Timeline:} $T_{total} = T_{App A} + T_{App B} + T_{App C}$

\subsubsection{Coarse-grained Parallelism}
The agent identifies independent sub-tasks at the application level.
It executes the price queries in App A and App B simultaneously.
The system blocks the execution of App C until both App A and App B have fully completed their tasks and returned the results.

\textit{Timeline:} $T_{total} = \max(T_{App A}, T_{App B}) + T_{App C}$

\subsubsection{Fine-grained Parallelism}
The agent exploits step-level parallelism by analyzing data dependencies.
While the price query sub-tasks in shopping apps (App A and App B) are being executed, the agent simultaneously operates in App C (e.g., searching for the contact, entering the chat interface, and activating the input field).
The execution in App C is suspended \textit{only} at the specific step where the message content (the prices) is required.
Once the query results from App A and App B become available, the execution in App C resumes immediately to send the message.

\textit{Timeline:} $T_{total} = \max(T_{App A}, T_{App B}, T_{App C\_setup}) + T_{App C\_send}$

\subsection{Real-World Application Coverage}

To ensure comprehensive evaluation across diverse usage patterns, we tested the system with 25 mainstream applications covering Social, Shopping, Information, and Video domains.
These applications represent the most commonly used mobile apps in real-world scenarios.

\subsubsection{Application Categories}

The tested applications span four functional categories corresponding to typical user activities.
Table~\ref{tab:app-categories} lists the applications in each category.

\begin{table}[b]
\setlength{\belowcaptionskip}{5pt}
\caption{Real-world applications tested across four functional categories.}
\centering
\footnotesize
\begin{tabular}{lp{5cm}}
\hline
\textbf{Category} & \textbf{Applications \& Description} \\
\hline
Social & WeChat, QQ, Sina Weibo (instant messaging \& social networking) \\
Shop & Taobao, JD.com, Pinduoduo, Xianyu, Ele.me, Meituan (e-commerce \& services) \\
Search & Xiaohongshu, Zhihu, Toutiao, Dianping, Browser (info retrieval \& tools) \\
Video & Bilibili, iQIYI, Tencent Video, Youku, Douyin, Kuaishou (video streaming) \\
\hline
\end{tabular}
\label{tab:app-categories}
\end{table}

\subsubsection{Task Scenarios}

We select 6 distinct task scenarios that represent common real-world multi-app workflows.
These scenarios range from simple sequential operations to complex parallel coordination tasks.
By varying the specific applications used within each category and adjusting task parameters, we create a total of 50 test instances covering diverse usage patterns.

Table~\ref{tab:task-scenarios} describes the task scenarios and provides example instructions.

\begin{table}[t]
\setlength{\belowcaptionskip}{5pt}
\centering
\footnotesize
\begin{tabular}{lp{4.5cm}}
\hline
\textbf{Scenario} & \textbf{Logic \& Example Instruction} \\
\hline
search+shop+social & \textbf{Serial/Pipeline:} Find recommended 2025 Canon cameras on Xiaohongshu, search on Taobao, send details via WeChat. \\
multi-video+social & \textbf{Parallel Query:} Check  iQIYI, Tencent Video for \textit{Joy of Life 3} updates, notify via WeChat if found. \\
multi-shop+social & \textbf{Parallel Comparison:} Query DJI Action 5 price on Taobao and JD.com, send comparison via WeChat. \\
single-shop+social & \textbf{Simple Pipeline:} Query DJI Action 5 price on Taobao, send result via WeChat. \\
search+social & \textbf{Info Sharing:} Search Disney Christmas event dates on Xiaohongshu, send schedule via WeChat. \\
search+shop & \textbf{Decision \& Action:} Find best-rated Sony headphones under 1000 CNY on Zhihu, order on Taobao. \\
\hline
\end{tabular}
\caption{Task scenarios representing real-world usage patterns.}
\label{tab:task-scenarios}
\end{table}

\subsubsection{Test Instance Construction}

To ensure the evaluation reflects realistic usage diversity, we construct the 50 test instances according to the following principles:

\textbf{Application Variation:} For each task scenario, we randomly selected compatible applications within each category to ensure the agent handles different UI designs and interaction patterns.
For example, shopping tasks alternated between Taobao, JD.com, and Pinduoduo.

\textbf{Parameter Diversity:} We varied search keywords, product names, target prices, and contact names across different instances to prevent result caching and ensure genuine task execution.

\textbf{Cross-Category Coordination:} We included tasks linking Video and Social apps (e.g., Douyin + QQ), Shopping and Search apps (e.g., Taobao + Zhihu), and Information Retrieval and Shopping apps (e.g., Xiaohongshu + Taobao) to test the scheduler's ability to handle data passing between diverse application architectures.

\subsection{Evaluation Methodology}

For each task instance, we execute the workflow using all three execution modes.
The primary metric is the end-to-end latency (seconds), measured from the moment the user issues the command until the final action (e.g., message sent or order placed) is confirmed by the system.
We calculate the speedup of the parallel modes relative to the serial baseline to quantify the efficiency improvements provided by the Agent Scheduler.
Each task instance is executed multiple times to ensure measurement reliability, and we report the average latency across runs.
	\nopagebreak

\end{sloppypar}
\end{document}